\documentclass{article}

\usepackage{arxiv}

\usepackage[utf8]{inputenc} 
\usepackage[T1]{fontenc}    

\usepackage[dvipsnames]{xcolor}
\usepackage{color}
\usepackage{colortbl}
\usepackage{hyperref}
\hypersetup{
    linkbordercolor = white,
    colorlinks=true,
    citecolor=Orchid,
    linkcolor=YellowOrange,
    urlcolor=RoyalPurple,
}

\usepackage{amsmath,amsfonts,bm}

\def\eqref#1{equation~\ref{#1}}
\def\Eqref#1{Equation~\ref{#1}}

\def\1{\bm{1}}

\def\rx{{\textnormal{x}}}
\def\ry{{\textnormal{y}}}
\def\rz{{\textnormal{z}}}

\def\vs{{\bm{s}}}

\def\vx{{\bm{x}}}
\def\vy{{\bm{y}}}
\def\vz{{\bm{z}}}

\def\evs{{s}}

\def\evy{{y}}
\def\evz{{z}}

\def\mX{{\bm{X}}}
\def\mY{{\bm{Y}}}

\DeclareMathAlphabet{\mathsfit}{\encodingdefault}{\sfdefault}{m}{sl}
\SetMathAlphabet{\mathsfit}{bold}{\encodingdefault}{\sfdefault}{bx}{n}

\def\gF{{\mathcal{F}}}
\def\gG{{\mathcal{G}}}
\def\gH{{\mathcal{H}}}

\def\sH{{\mathbb{H}}}

\def\sL{{\mathbb{L}}}

\def\sS{{\mathbb{S}}}

\def\sV{{\mathbb{V}}}

\usepackage{url}            
\usepackage{nicefrac}       
\usepackage{microtype}      
\usepackage{lipsum}		
\usepackage{graphicx}
\usepackage{doi}

\usepackage{cite}

\usepackage{subcaption}
\usepackage{array}
\usepackage{graphbox}

\usepackage{longtable}
\usepackage{tabularray}
\usepackage{rotating}
\usepackage{tabularx}
\usepackage{booktabs}
\usepackage{multicol}
\usepackage{multirow}
\usepackage{diagbox}

\usepackage[safe]{tipa} 
\usepackage{tikz}  
\newcommand{\graphZero}{%
\begin{tikzpicture}[scale=0.4, baseline=(current bounding box.center)]
  \draw (0,0) -- (1,0) -- (0.5,0.866);
  \draw[fill=black] (0,0) circle (2.0pt);
  \draw[fill=black] (1,0) circle (2.0pt);
  \draw[fill=black] (0.5,0.866) circle (2.0pt);
\end{tikzpicture}%
}

\newcommand{\graphOne}{%
\begin{tikzpicture}[scale=0.4, baseline=(current bounding box.center)]
  \draw (0,0) -- (1,0) -- (0.5,0.866) -- cycle;
  \draw[fill=black] (0,0) circle (2.0pt);
  \draw[fill=black] (1,0) circle (2.0pt);
  \draw[fill=black] (0.5,0.866) circle (2.0pt);
\end{tikzpicture}%
}

\newcommand{\graphTwo}{%
\begin{tikzpicture}[scale=0.4, baseline=(current bounding box.center)]
  \draw (0,0) -- (1,0) -- (1,1) -- (0,1);
  \draw[fill=black] (0,0) circle (2.0pt);
  \draw[fill=black] (1,0) circle (2.0pt);
  \draw[fill=black] (1,1) circle (2.0pt);
  \draw[fill=black] (0,1) circle (2.0pt);
\end{tikzpicture}%
}

\newcommand{\graphThree}{%
\begin{tikzpicture}[scale=0.4, baseline=(current bounding box.center)]
  \draw (0,0) -- (1,0) -- (1,1) -- (0,1) -- cycle;
  \draw[fill=black] (0,0) circle (2.0pt);
  \draw[fill=black] (1,0) circle (2.0pt);
  \draw[fill=black] (1,1) circle (2.0pt);
  \draw[fill=black] (0,1) circle (2.0pt);
\end{tikzpicture}%
}

\newcommand{\graphFour}{%
\begin{tikzpicture}[scale=0.4, baseline=(current bounding box.center)]
  \draw (1,1) -- (1,0);
  \draw (1,1) -- (0,1);
  \draw (1,1) -- (0,0);
  \draw[fill=black] (0,0) circle (2.0pt);
  \draw[fill=black] (1,0) circle (2.0pt);
  \draw[fill=black] (0,1) circle (2.0pt);
  \draw[fill=black] (1,1) circle (2.0pt);
\end{tikzpicture}%
}

\newcommand{\graphFive}{%
\begin{tikzpicture}[scale=0.4, baseline=(current bounding box.center)]
  \draw (0,0) -- (1,1);
  \draw (0,0) -- (1,0);
  \draw (0,0) -- (0,1);
  \draw (1,1) -- (1,0);
  \draw[fill=black] (0,0) circle (2.0pt);
  \draw[fill=black] (0,1) circle (2.0pt);
  \draw[fill=black] (1,0) circle (2.0pt);
  \draw[fill=black] (1,1) circle (2.0pt);
\end{tikzpicture}%
}

\newcommand{\graphSix}{%
\begin{tikzpicture}[scale=0.4, baseline=(current bounding box.center)]
  \draw (0,0) -- (1,0) -- (1,1) -- (0,1) -- cycle;
  \draw (0,0) -- (1,1);
  \draw[fill=black] (0,0) circle (2.0pt);
  \draw[fill=black] (1,0) circle (2.0pt);
  \draw[fill=black] (1,1) circle (2.0pt);
  \draw[fill=black] (0,1) circle (2.0pt);
\end{tikzpicture}%
}

\newcommand{\graphSeven}{%
\begin{tikzpicture}[scale=0.4, baseline=(current bounding box.center)]
  \draw (0,0) -- (1,0) -- (1,1) -- (0,1) -- cycle;
  \draw (0,0) -- (1,1);
  \draw (1,0) -- (0,1);
  \draw[fill=black] (0,0) circle (2.0pt);
  \draw[fill=black] (1,0) circle (2.0pt);
  \draw[fill=black] (1,1) circle (2.0pt);
  \draw[fill=black] (0,1) circle (2.0pt);
\end{tikzpicture}%
}

\title{Studying and Improving Graph Neural Network-based Motif Estimation}

\date{} 					

\author{
    Pedro C.~Vieira
        \thanks{
        Correspondence to: pedrocvieira@fc.up.pt} \\
	Department of Computer Science\\
	Faculty of Science of the University of Porto\\
	Porto, Portugal \\
	\texttt{pedrocvieira@fc.up.pt} \\
\And
Miguel E.~P.~Silva \\ 
	Department of Computer Science\\
	Faculty of Science of the University of Porto\\
	Porto, Portugal \\
	\texttt{mepsilva@fc.up.pt} \\
\And
Pedro Manuel Pinto Ribeiro \\ 
	Department of Computer Science\\
	Faculty of Science of the University of Porto \\ 
    CRACS/INESC-TEC\\
	Porto, Portugal \\
	\texttt{pribeiro@fc.up.pt} \\
}

\renewcommand{\shorttitle}{Studying and Improving GNN-based Motif Estimation}

\hypersetup{
pdftitle={Studying and Improving Graph Neural Network-based Motif Estimation},
pdfsubject={cs.AI, cs.LG},
pdfauthor={Pedro C.~Vieira, Miguel E.~P.~Silva, Pedro Manuel Pinto Ribeiro},
pdfkeywords={Graph Neural Networks. Motifs, Significance-Profile},
}

\begin{document}
\maketitle

\begin{abstract}
	Graph Neural Networks (GNNs) are a predominant method for graph representation learning. However, beyond subgraph frequency estimation, their application to network motif significance-profile (SP) prediction remains under-explored, with no established benchmarks in the literature. We propose to address this problem, framing SP estimation as a task independent of subgraph frequency estimation. Our approach shifts from frequency counting to direct SP estimation and modulates the problem as multitarget regression. The reformulation is optimised for interpretability, stability and scalability on large graphs. We validate our method using a large synthetic dataset and further test it on real-world graphs. Our experiments reveal that 1-WL limited models struggle to make precise estimations of SPs. However, they can generalise to approximate the graph generation processes of networks by comparing their predicted SP with the ones originating from synthetic generators. This first study on GNN-based motif estimation also hints at how using direct SP estimation can help go past the theoretical limitations that motif estimation faces when performed through subgraph counting.
\end{abstract}


\section{Introduction}\label{sec:intro}

In this work, ``graph'' and ``network'' are used interchangeably. A graph is deemed a network motif~\cite{Milo2002} when it occurs more frequently than random chance would suggest, serving as a powerful tool for analysing complex networks. Recognising significant substructures helps elucidate a graph's underlying organizational principles, advancing both theoretical and practical understanding, especially in biology. For example, the feed-forward loop is a key functional pattern in gene regulation networks \cite{Mangan2003}. It has also been discovered that motifs enable efficient communication and fault-tolerance across transcriptional networks \cite{Roy2020}.

Discovering a motif entails counting the number of occurrences of the desired structure, both in the network in study and in a set of control networks (networks that preserve key properties of the original) to understand its significance. This process is at least NP-complete, as it includes the subtask of determining whether a subgraph exists within a larger network (the subgraph isomorphism problem). Although there are methods to perform an analysis based on motifs, \cite{Ribeiro2021}, they have a high temporal complexity, rendering them intractable for very large networks. 

A good alternative that can help overcome these limitations is to use machine learning, specifically Graph Neural Networks (GNNs), to perform motif estimation. However, as far as we are aware, there are few methods that attempt this route. Existing methods \cite{Zhang2020-micro, Oliver2022,Ying2019-gnexplainer,Chen2023a-tempexplainer,Chen2023-motif-graph-neural,Sheng2024-MGAT,Dareddy2019-motif2vec,lee2018-high-gcn,Fu2023a-DeSCo} typically adopt formulations that are too complex and lead to the omission of important aspects in traditional motif estimation, deviating significantly from the original concept of motif~\cite{Milo2002, Milo2004-SuperFamilies}. 
More importantly, we postulate that using GNNs just as a subgraph frequency estimation method to then use does estimations for calculating motif scores fundamentally misses the opportunity to exploit GNNs for more powerful and direct formulations. 
We identify that using GNNs as subgraph frequency estimators limits the task of motif estimation to the expressivity that a model can achieve at subgraph counting. In addition to this, we find that other motif estimation methods \cite{Zhang2020-micro, Oliver2022} only provide an indication regarding which subgraphs are motifs, whereas real-world scenarios require more detailed information on the significance of the found motifs, information regarding which subgraphs are not motifs and sometimes information regarding anti-motifs.

In contrast, \textbf{our work} is aimed at addressing these shortcomings by making small but fundamental changes. Specifically, we propose (1) estimation of motif scores that are interpretable and can be used in comparisons between networks and other downstream tasks; (2) clear statistical relevance with connections to traditional estimation based on a null model; (3) control over what graphs will be evaluated as candidate motifs; (4) complete description of the chosen graphs, allowing a detailed analysis of how a graph is placed in the space of motifs; (5) stability and time complexity improvements when estimating out-of-distribution with respect to the network size; and (6) allowing the formulation to possibly bend expressivity results for motif estimation based on subgraph counting.

\textbf{Key Contributions:} Our key contributions can be summarised as follows:  
\textbf{{\textit{(1)}}} We show preliminary results that the difficulty of motif discovery with Message Passing Neural Networks (MPNNs) can be manipulated through different formulations of the target variable (e.g. different concept of what is a motif). Hence, depending on the formulation used, motif estimation does not have to follow the limitations regarding subgraph counting exposed in the literature. \textbf{{\textit{(2)}}} We present a formulation -- multi-target regression of normalized significance-profiles -- based on simple assumptions that improves the ability of MPNNs to perform motif estimation while approximating this task to traditional motif estimation; \textbf{{\textit{(3)}}} We make available a large and diverse synthetic dataset in terms of both graph topology and motif significance-profile, using twenty three random network models. We also present a collection of more than 100 real-world networks and their motif significance-profile. 

\section{Background and Related Work}\label{sec:relat-work}

Let a graph $\gG = (\sV_\gG, \sL_\gG, \mX)$ where $\sV_\gG$ denotes the vertex set of $\gG$, $\sL_\gG \subseteq \sV_\gG \times \sV_\gG$ the edge and $\mX \in \mathbb{R}^{d_1 \times d_2}$ the vertex features such that $\forall v \in \sV_\gG, \vx \in \mathbb{R}^{d_2}$. Let all edges be undirected such $(u,v) \in \sL_\gG \Leftrightarrow (v,u) \in \sL_\gG$. 
Let $\gH$ be a subgraph of $\gG$ if and only if $\sH_\gH \subseteq \sV_\gG \land \sL_\gH \subseteq \sL_\gG$, such that exists an injective homomorphism given by the injective function $f: \sV_\gH \mapsto \sV_\gG$ yielding $(v,u) \in \sL_\gH \Rightarrow (f(v),f(u)) \in \sL_\gG$. If $f$ is bijective and $f^{-1}$ is an homomorphism (injective by construction) the relation is an isomorphism and the subgraph induced.

In order to discover motifs, we must define three steps: (1) What is the set of graphs, $S_G$, that we admit as candidates for motifs; (2) What method is used to count the occurrences of graphs of $S_G$ in the graph of interest $\gG$; (3) How is the significance of the obtained counts calculated.

\subsection{Step One - How to Define the Set of Graphs Used}\label{sub:motif-step1}
In this step, $S_G$ is typically defined \textit{a priori}. This method is the most widely used,~\cite{Milo2002, Milo2004-SuperFamilies, Shen-Orr2002}, and in most common cases, the selection of graphs used are ones known to be important to the area of the work in question~\cite{Shen-Orr2002, Alon2007}.

Defining $S_G$ \textit{a priori} is frequent for ``non-machine-learning'' techniques, but it is also common in machine-learning ones~\cite{rong2020-grover, Ying2020-SPMiner}. However, when using techniques based on machine-learning, it is easier to create a task that can infer structures in $\gG$ than when using non-machine learning approaches. Hence, motif discovery can be modulated as the task of finding the best set of graphs that can be considered motifs. To achieve this, it is typically used a graph metric (not necessarily in the formal metric sense) to measure how much a motif a graph is.
For example, using maximum-likelihood estimation to attribute subgraph embeddings to certain graphs or by devising loss functions that penalise agreement with control networks~\cite{Zhang2020-micro, Oliver2022}. 

\subsection{Step Two - How to Count Subgraphs}\label{sub:motif-step2}
\textbf{Non-GNN Methods.} Numerous methods exist for approximating subgraph counts, eschewing GNNs or any machine learning techniques. We refer the interested reader to~\cite{Ribeiro2021} for a survey of these methods.

\textbf{GNN Methods.} Counting occurrences of a graph $\gG$ in another graph $\gH$ using GNNs was first introduced by Chen et al.~\cite{Chen2020b}. This work introduced significant limitations of what substructures MPNNs can count. Subsequent works have refined MPNN-like models to be more expressive, allowing them to have guarantees of being able to count occurrences of more graphs. One branch of such models is known as node-rooted Subgraph GNNs. These will extract a $k$-hop neighbourhood for each node $v$ in the graph to be studied and calculate an added feature for $v$.
These architectures, with models as powerful as 1-WL as the backbone, are strictly more powerful than maximum powerful MPNNs but are less powerful than the 3-WL test~\cite{Frasca2022a,Yan2023a,Zhang2023-subwl}. Hence, they have limitations regarding the type of structures that they can count. Huang et al.~\cite{Huang2022b} gives a characterisation of what substructures node-rooted Subgraph GNNs cannot count at node-level. 
They show that the Subgraph GNNs cannot count cycles of more than four and paths of more than three nodes.


Recently a new theoretical view of Subgraph GNNs based on the Subgraph Weisfeiler-Lehman, a new version of the WL test, has been proposed~\cite{Zhang2023-subwl}. This analysis presents a characterisation of the expressive power of all node-rooted Subgraph GNNs. They conclude that no node-rooted Subgraph GNN can be more powerful than the 2-folklore-WL (3-WL) (also noticed in~\cite{Yan2023a,Frasca2022a}). In that same work, the authors demonstrate that no node-rooted Subgraph GNN can achieve the maximum expressivity of their time complexity class. This result draws a limitation in the design of node-rooted Subgraph GNNs. Regarding induced subgraph counting at graph level, the subject of our work, 3-WL models can count all patterns of $3$ or less nodes~\cite{Lanzinger2023} and 1-WL models cannot count any pattern with $3$ or more nodes.

\subsection{Step Three - How is Significance Obtained}\label{sub:motif-step3}
After obtaining the frequency of the structures in $S_G$, the next step is to evaluate their significance. Hence, it is necessary to have an idea of what would produce, with no factor other than random chance, a (control) network similar to $\gG$ for some characteristic of interest. Let us denote as NULL a model that can achieve that goal. One example of NULL is a model that, given a graph $\gG$, randomly switches edges while keeping the degree distribution of $\gG$ -- degree distribution is the characteristic of interest. The rewiring process is completely random and without any bias towards any predisposition~\cite{Milo2002, Milo2004-SuperFamilies}. A myriad of NULL models can be conceived depending on the characteristic to studied.

\subsection{Motifs and Graph Neural Networks}\label{sub:rel-motif}

Motif estimation, when approached through the lens of GNNs, appears to be a challenge that, to the best of our knowledge, remains largely unexplored in the existing literature. Despite using the term ``motif'', works like~\cite{besta-motifs-seam} do not overlap with our work. They attend to the identification of the chance of higher-order structures to appear and they can be seen as a generalisation of link prediction.
We will now outline the interactions that we have identified. 

\textbf{Directly counting.}
One of the approaches that better matches direct motif estimation with GNNs involves two main steps. First, a GNN, such as a node-rooted Subgraph GNN, is used to count the occurrences of the graphs of interest, $S_G$, within the input graph, $\gG$. Second, a suitable null model is selected, and $T$ graphs are generated according to this model. The same GNN used for $\gG$ is then applied to count the occurrences of $S_G$ in each of the $T$ control graphs. Additionally to generic GNNs, SPMiner~\cite{Ying2020-SPMiner} is an example of a method that specialises in subgraph counting.

\textbf{Motifs as tool.}
Other works that integrate GNNs and motifs typically deviate from estimating motifs and use pre-computed ones to enhance the power of GNNs or explain decisions. Examples of this work include Motif Convolutional Networks~\cite{lee2018-high-gcn}, motif2vec~\cite{Dareddy2019-motif2vec}, Motif Graph Attention Network~\cite{Sheng2024-MGAT}, Motif Graph Neural Network~\cite{Chen2023-motif-graph-neural}, Heterogeneous Motif Graph Neural Network~\cite{Yu2022-HM-GNN}, GNNExplainer~\cite{Ying2019-gnexplainer} and TempME~\cite{Chen2023a-tempexplainer}.

\textbf{Learning Motifs.} 
The works that directly address motif estimation, such as those that later used them to refine downstream tasks, include MICRO-Graph~\cite{Zhang2020-micro} and MotiFiesta~\cite{Oliver2022}. 

We will ignore the methods that use motif as a tool since they reside out of the scope of the problem. The main problems of the presented works are the following: (1) either the model does not assume a null model and returns raw counts of occurrences of a general $\gH$ in $\gG$ (SPMiner and other frequency estimation models), or (2) the model may use a null model to guide motif search but only returns the subgraph(s) that are considered a motif by the model, meaning it is typically not possible to query for a specific $\gH$ (MICRO-Graph and MotiFiesta). In fact, methods in this category commonly encounter challenges in regulating the size and shape of the graph proffered as a candidate motif.

Additionally, models that return the raw count of occurrences (case (1)) can suffer from poor generalisation since the number of graph structures grows super-exponentially~\cite{Fu2023a-DeSCo}. Hence, as the size of a graph $\gG$ grows, the possible counts of a substructure $\gH$ in $\gG$ also grow super-exponentially, causing high variation between results of small and very-large graphs when estimating out-of-distribution with respect to the network size. This fact can hinder the learning process of models that aim at being agnostic of network size and topology. Furthermore, since these methods essentially perform subgraph counting, they are bounded by the limitations of subgraph counting. This means that the task of motif estimation is limited by the expressivity that a model can achieve at subgraph counting.
Finally, for raw count models, since no null model is assumed, obtaining significance implies subsequent computation.

As for models falling under category (2), they typically ignore everything not branded as a motif, sometimes not even returning a motif score for the graphs regarded as such.
Additionally, since the metric employed for discerning the suitable graph for a candidate motif has to be a learnable function, it can be hard to interpret its meaning in order to evaluate the strength of the motif. This can lead to the lack of clarity in the application to real-world scenarios in downstream tasks e.g. compare strength across networks. Finally, if no null model is applied, like in the case of category (1) obtaining significance implies subsequent computation.



\section{Method}\label{sec:method}

Hereafter, as per the first paragraph of Section \ref{sec:relat-work}, referencing the number of occurrences of a graph $\gH$ within $\gG$, denotes the induced count of $\gH$ in $\gG$. Furthermore, all graphs are undirected and they do not have edge features.

According to the definition of motif adopted, to understand if a graph $\gH$ is a motif of a graph $\gG$, we must know the number of occurrences of $\gH$ in $\gG$. 
Let us denote such count as $C(\gH, \gG)$. 
Furthermore, to grasp the importance of $\gH$ in $\gG$, it is needed to know the count of $\gH$ across sufficient graphs derived from a null model denoted as NULL (control graphs). The chosen null model is the one described in Section \ref{sub:motif-step3}, and uses the degree distribution as the characteristic of interest that will be used in the control networks~\cite{Milo2002, Milo2004-SuperFamilies}.
Let us denote the average count of $\gH$ in graphs derived from NULL as $C^{\mu}(\gH,\gG_\text{NULL})$ and the standard deviation as $C^{\sigma}(\gH,\gG_\text{NULL})$. 
Hence, $Z(\gH,\gG_\text{NULL}) = \frac{C(\gH,\gG)-C^{\mu}(\gH,\gG_\text{NULL})}{C^\sigma(\gH, \gG_\text{NULL})}$ denotes the standardization (Z-score) of the occurrences of a graph $\gH$ in $\gG$.



\subsection{Our Approach}\label{sub:our-approach} 
We first impose a restriction on the number and type of graphs the model will predict. We refer to the set of graphs used as $\Omega$. The function notation $\Omega(\gG)$ gives the set of all graphs that have the same number of connected nodes as the graph $\gG$.
The restriction of the number of graphs implies that the proposed model will not be able to search if an arbitrary graph is or is not a motif. However, by having a model that has a more restricted objective, we aim to achieve higher precision in the said objective while being able to fully understand how the graphs in $\Omega$ are positioned in the motif spectrum.

Furthermore, we do not follow the approach of predicting $C(\gH, \gG)$. Instead, we aim at directly modelling a statistical motif score like $Z(\gH, \gG)$. This achieves full transparency on what null model is used and how it is used. Additionally, this eliminates the need to compute multiple networks based on the null model to determine significance. In fact, this step is completely skipped, gaining a lot in performance, while the result remains statistically interpretable.

Instead of modelling the learning task as predicting a single value $Z(\gH, \gG)$ for some $\gH$ and some $\gG$, we model it as a multi-target regression problem in order to predict the motif score of multiple subgraphs at once. This formulation means that a model will train to predict all the graphs of $\Omega$ at the same time. Hence, this characterisation naturally allows the construction of motif profiles~\cite{Milo2004-SuperFamilies}.
Thus, we define a vector of Z-scores, $\vz = [Z(\gH_1, \gG) \dots Z(\gH_n, \gG)]$. 
Since $\Omega$ has a restricted size, one aspect that deserves careful consideration is deciding what is the size of $\Omega$ and what graphs compose it. 
Should the selected graphs exhibit negligible relation, an attempt to predict the Z-score concurrently for all graphs may prove harmful to the performance of the model. In this case, such an approach forces the model to incorporate distinct patterns to predict scores for each graph, thereby resulting in a sub-optimal global predictive efficacy.
However, if $\Omega$ is composed of a well-thought group of graphs, allowing them to share common patterns from a learning perspective, we hypothesise that the performance of the model can improve when compared to predicting just one Z-score, due to the possibility of what is learned about a target variable be ``shared'' with others through weight sharing (one other advantage is the need to only train a single model instead of multiple). A good candidate for $\Omega$ should have patterns that are interconnected with each other, either from the point of view of the Z-score distribution or from a topological one. 

Building on top of what was described in the last paragraph, we focus on small graphs, in particular all connected graphs of size three and four. This is also supported by existing relevant literature \cite{Milo2004-SuperFamilies, Milo2002, Shen-Orr2002, Asikainen2020, Przulj2007, Ribeiro2013} suggesting that to understand a complex network, it is important to understand how small graphs behave. We focus on these graphs because their proximity in size should allow them to have a topological connection that translates in a connection in their Z-scores. 
Restricting the size of the graphs used in $\Omega$ to small ones also has the added benefit that we can get the ground-truth of motif scores for a diverse type and size of networks, allowing for a richer train dataset. 
Furthermore, we expect that using a set of graphs of increasing size in the number of nodes and edges gives enough interconnectivity between their patterns from both a topological and a Z-score distribution point of view to allow the model to have a strong inductive bias towards meaningful patterns, allowing for a stronger performance. For example, a graph with many size four cliques will probably have a small amount of 4-stars.
For the chosen graphs it is possible to create two groups in $\Omega$, the graphs of size three and the graphs of size four. %

Finally, we normalise the Z-score, $Z(\gH, \gG)$, across groups of graphs, according to $\evs_i = {\evz_i}/{(\sum_{j \in \Omega(i)}\evz_j^2})^{1/2}$ ~\cite{Milo2004-SuperFamilies}. After normalisation, the values of $\vz$ are constrained between $-1$ and $1$, independent of network size. This will be beneficial at maintaining the predictive stability of the model when estimation out-of-distribution with respect to the network size. Additionally, this will further enhance the usage of the score for downstream tasks as it allows comparison across networks of different sizes.
In fact, this normalisation keeps a wide motif spectrum, allowing to make fine distinctions between graphs. For example, a graph with score $0.8$ appears more often that expected than a graph with motif score $0.6$.
Moreover, this normalisation imposes a mathematical interconnectivity between the Z-scores of graphs of the same group. This relationship, where the sum of squared normalised Z-scores equals $1$, supports a multi-target objective and further strengthens the problem formulation by adding an additional layer of interdependence among graphs.
Let us denote the normalised Z-scores as, $\vs$, also known as significance-profiles. The learning task thus consists of minimising the MSE between the true and predicted significance-profiles.



\subsection{On the Relation with Expressivity Through Subgraph Counting}\label{sub:sp-vs-count}

It is expected that the expressivity regarding substructure counting to be highly related to the expressivity of discovering the significance-profile of graphs. Concretely, the problem of counting graphs is a subset of the problem of discovering significance-profiles where reducing the null model to nothing reduces the problem to graph counting. 

Since $P$, the problem of counting graphs, is a subset of $S$, the problem of significance-profile estimation, it is possible to obtain instances of $S$ that are as hard as $P$, easier than $P$ and harder than $P$. Under the assumption that $S$ and $P$ function around the same set of graphs, these differences in difficulty come from the choice of null model. In the case of $S = P$, the null model should do nothing, for example, returning always $0$. In the case of $S < P$, the null model could always return the counts of each subgraph in a graph $\gG$ without modifying $\gG$, reducing the problem to always predicting a vector of zeros. For the case of $S > P$, employing a null model that randomly returns counts for $\gG$ should make the problem theoretically harder since the model would have to learn the random process employed to correctly construct the significance-profiles.
Thus, theoretical guarantees of expressivity might not hold depending on the selected null model. For instance, a recent demonstration solved the dimensionality of the $k$-WL test for induced subgraph count. It is stated that to perform induced subgraph count of any pattern with $k$ nodes we need at least dimensionality $k$~\cite{Lanzinger2023}. Furthermore, no induced pattern with $k+1$ nodes can be counted with dimensionality $k$, a result not verified to non-induced counts~\cite{Lanzinger2023}.
However, when working with significance-profiles over graphs of size $k$, the relation is more complex since it depends on the null model.


\textbf{Testing with 1-WL bounded models?} MPNNs cannot perform induced counts of patterns of three or more nodes~\cite{Chen2020b}. Nevertheless, MPNNs are not inherently incapable of counting patterns in any graphs. Rather, for a pattern $\gH$, there exists graphs $\gG_1, \gG_2 \in \mathcal{\gG}$ such that $C(\gH, \gG_1) \neq C(\gH, \gG_2)$ and for any MPNN $M$ under 1-WL, $M(\gG_1) = M(\gG_2)$.
Hence, $M$ cannot discover $C(\gH, \gG_1)$ and $C(\gH, \gG_2)$ simultaneously. 
However, within the 1-WL framework, MPNNs remain valuable and find practical applications in real-world scenarios. Furthermore, we did not construct any characterisation of the problem space of $S$ regarding $P$ for the null model used. Hence, we might have made the problem easier (or harder) than substructure counting. 
Thus, we will scrutinise the capability of MPNNs to address our particular challenge. Furthermore, testing with more expressive models like Subgraph GNNs is hard due to their complexity~\cite{Frasca2022a}.


\section{Datasets}\label{sec:data}

We are not interested in limit testing the power of our formulation or comparing it to theoretical tests: rather, our goal is to test our model on a large dataset with diverse topological features and motif scores. Hence, we avoid standard GNN datasets~\cite{wu2018moleculenet, morris2020tudataset, hu2021open, hu2021ogblsc}. 
Given the difficulty of obtaining a real-world dataset spanning multiple domains while retaining the defined properties, we rely exclusively on synthetic graphs. Concretely, we use 23 synthetic generators (11 non-deterministic, 12 deterministic). We explore their graph-generating space to extract all types of topologies while limiting the graph size to avoid excessive training times. 
The final dataset contains $109164$ graphs in the non-deterministic segment and $38400$ in the deterministic ($\approx$ 250 million nodes and $\approx$ 750 million edges). Finally, for the ground-truth, we calculate $\vs$ using G-Tries~\cite{Ribeiro2013}. 

Using synthetic data to train GNNs is not a new concept, but the most of the popularly used datasets typically have very small graphs (at most few hundreds of nodes) and are generated from a small set of generators, often random regular graphs and Erd\H{o}s-Renyi graphs~\cite{Chen2020b}. Another popular type of synthetic graph dataset for benchmarking is small handcrafted graphs to limit test GNN models~\cite{Abboud2021, Murphy2019, pmlr-v139-balcilar21a, Wang2023}. While still very limited, the only exceptions identified use some Barabási-Albert graphs, some graphs with crude community structure, trees and grids~\cite{Velickovic2020-neural-execution, Corso2020}. In contrast, our approach employs multiple graph generators that simulate real-world phenomena, yielding a dataset with high topological diversity and a close resemblance to real data. 



Our analysis of the SPs for all graphs in the synthetic data reveals that the 3-path and triangle exhibit limited motif scores, namely $\{\pm1/\sqrt{2},0,1\}$ for the 3-path and $\{\pm1/\sqrt{2},0\}$ for the triangle. 
Consequently, their Z-scores are interdependent. 
In fact, except when both are zero or when the 3-path is $1$ and the triangle is $0$ (an artefact of the G-Trie model), they are symmetric. 
Additionally, the significance profile of size-four graphs is partly encoded in the size-three profile, suggesting an advantage in using both sizes as target variables. 
For only size-four, we observed no further strict dependencies other than the mathematical constraint stated in Section~\ref{sec:method}.

\textbf{Real-World Data.} Since we are interested in assessing the performance of the models with real-world data, we compiled a dataset based on real networks of multiple categories. 
Besides varying the type of network, we vary in their relative size in terms of number of nodes and edges. 
We have categorised them into two groups based on their average size with respect to the training set: (1) small-scale networks, which range from slightly smaller to around average the size of synthetic ones, and (2) medium-large-scale, which exceed (sometimes significantly) the average synthetic network size.

The additional material (Section~\ref{sec:data-details}) provides more details for this section.



\section{Methodology}\label{sec:methodology}

The model used in the experiments is similar to the one described in by Chen et. al \cite{Chen2020b}, definition A.1. from Appendix A. We utilise the same architecture only modifying the target and loss to the ones described in Section~\ref{sec:method}. 
We employ GIN~\cite{Xu2019}, GAT~\cite{Velickovic2018}, GraphSage~\cite{Hamilton2017SAGE} and GCN~\cite{Kipf2017} as backbones and optimise all models with Optuna so that they can represent the best possible result for the used parameter space. We train two instances, one based on non-deterministic (ND) datasets and other on deterministic (D) ones.

\textbf{``Correct'' Predictions.} 
Criteria for a prediction of a significance-profile to be correct/useful depends largely on the research field. Hence, we evaluate the result on multiple simple error thresholds, namely, $\leq5\%, \leq10\%, \leq25\%$ and $<50\%$. The $50\%$ threshold means that all predictions that match just the signal are considered as correct.
We count predictions divided by synthetic generator. Furthermore, we ensure that for a prediction to count as ``correct'', SP values for all graphs in $\Omega$ must have the correct sign. Hence, for example, for the $\leq5\%$ threshold, it is not enough for the error to be below $2\times0.05 = 0.1$, where $2$ is maximum possible error, the signal of the predictions must also match.
 

Let $T$, be a naive benchmark model, predicting a random significance-profile, but taking into account the restrictions, described in Section~\ref{sec:method}, on the range of values each group of $\Omega$ can take.
For the defined thresholds, according to $1e6$ simulations, $T$ will have a rate of ``correct'' guesses of $0.0001\%$ for $5\%$, $0.0019\%$ for $10\%$ $0.1108\%$ for $25\%$ and $0.4012\%$ for $50\%$.

The additional material (Sections~\ref{sec:methdology-details} and~\ref{sec:exp-res-appx}) presents more details about the model, how it was trained and the results for more thresholds. \href{https://github.com/PedrV/motifs-estimation-with-gnns}{Github link} for code and replication steps.












\section{Results}\label{sec:results}

For the ND segment, the best models were an instance of GIN and one of GraphSage (SAGE). As for the D segment, an instance of GIN stood as the undisputed best. All others were significantly worse and hence not further analysed.

\subsection{Predictions in the Synthetic Dataset}\label{sub:model-predictions-synt}

\begin{figure}[htbp]
  \centering
  \begin{subfigure}[t]{0.23\textwidth}
    \includegraphics[width=1.1\linewidth]{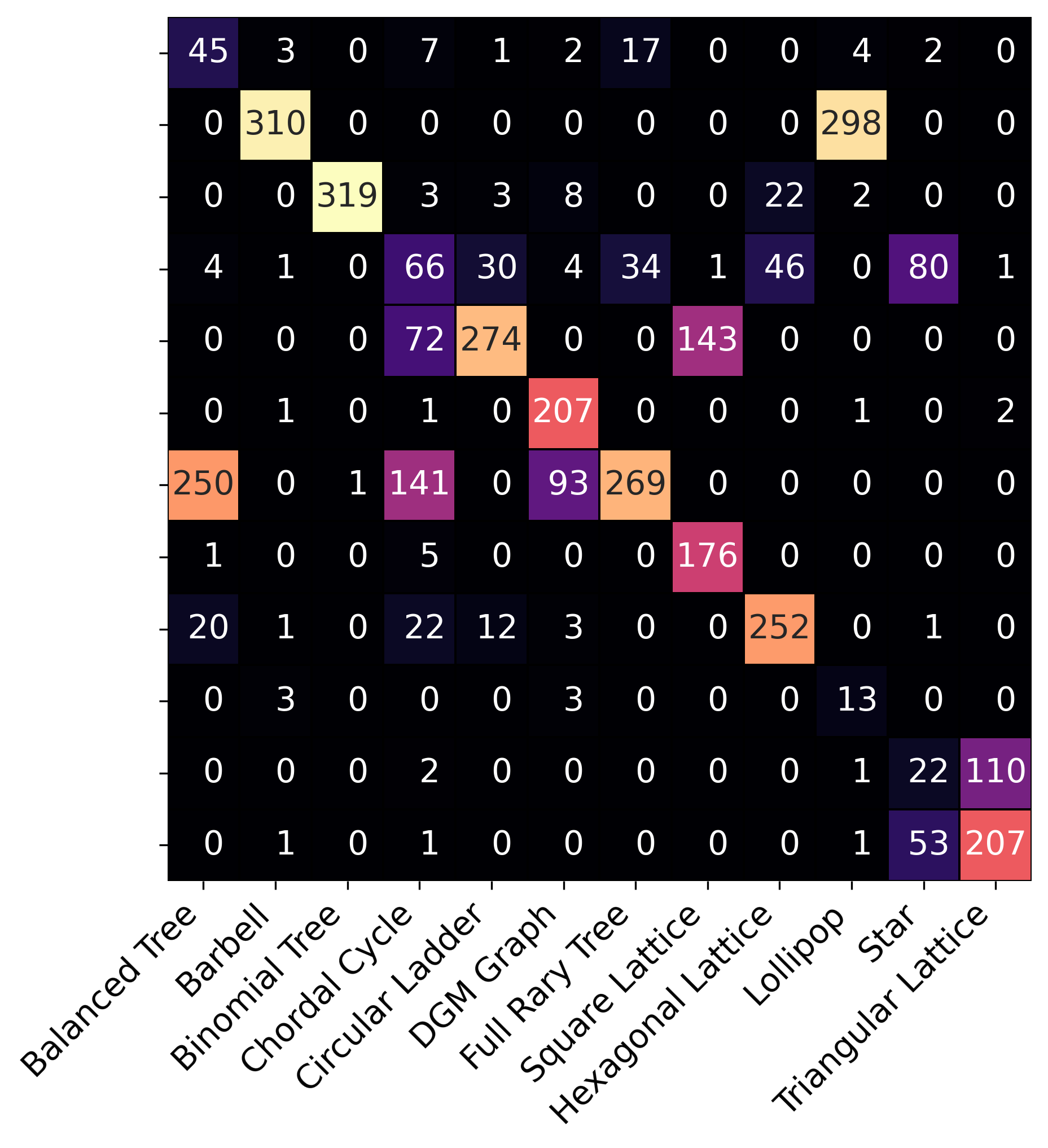}
    \caption{GIN D}
    \label{fig:dgin}
  \end{subfigure}\hspace{0.02\textwidth}%
  \begin{subfigure}[t]{0.23\textwidth}
    \includegraphics[width=1.1\linewidth]{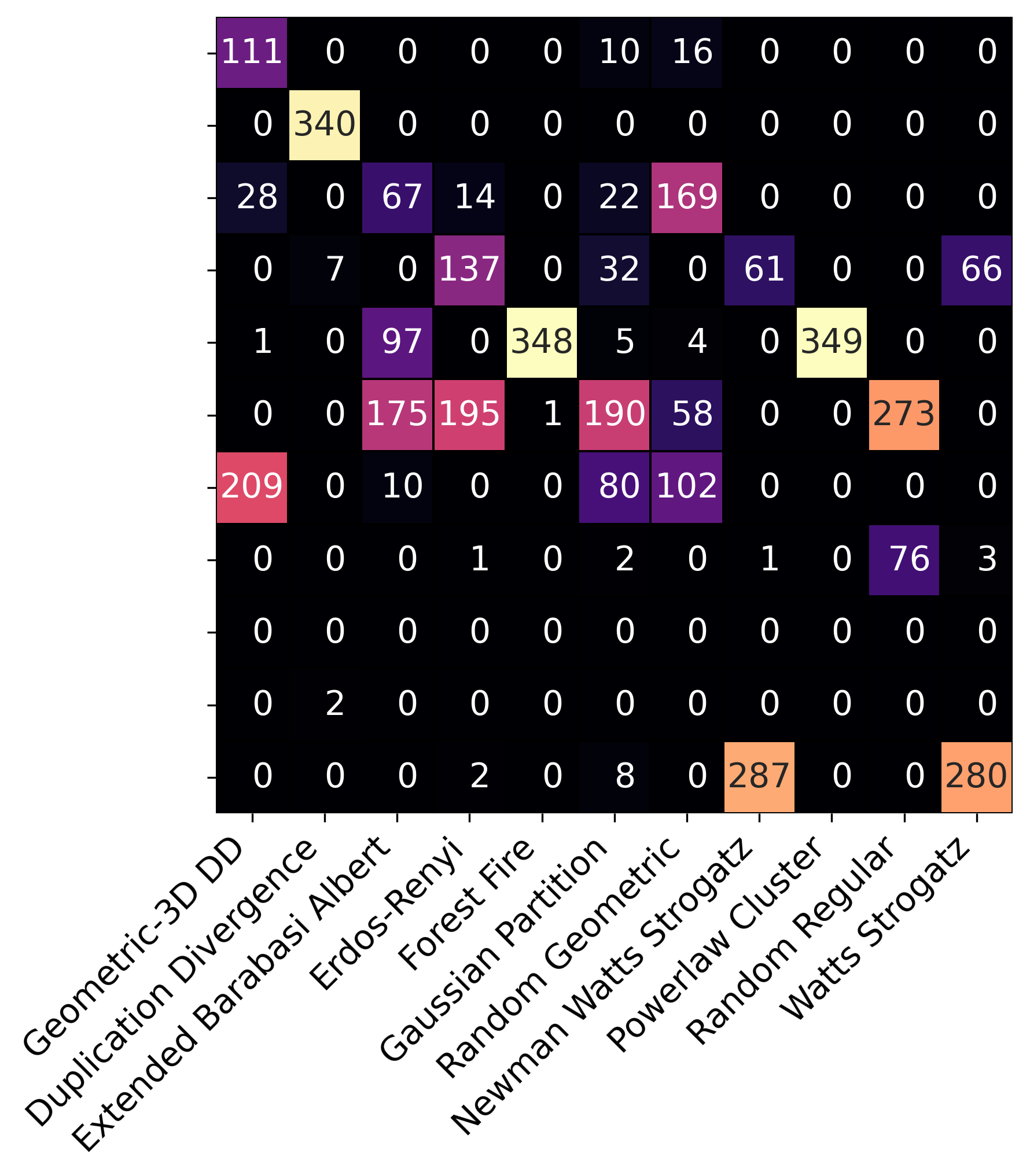}
    \caption{GIN ND}
    \label{fig:ndgin}
  \end{subfigure}\hspace{0.02\textwidth}%
  \begin{subfigure}[t]{0.23\textwidth}
    \includegraphics[width=1.1\linewidth]{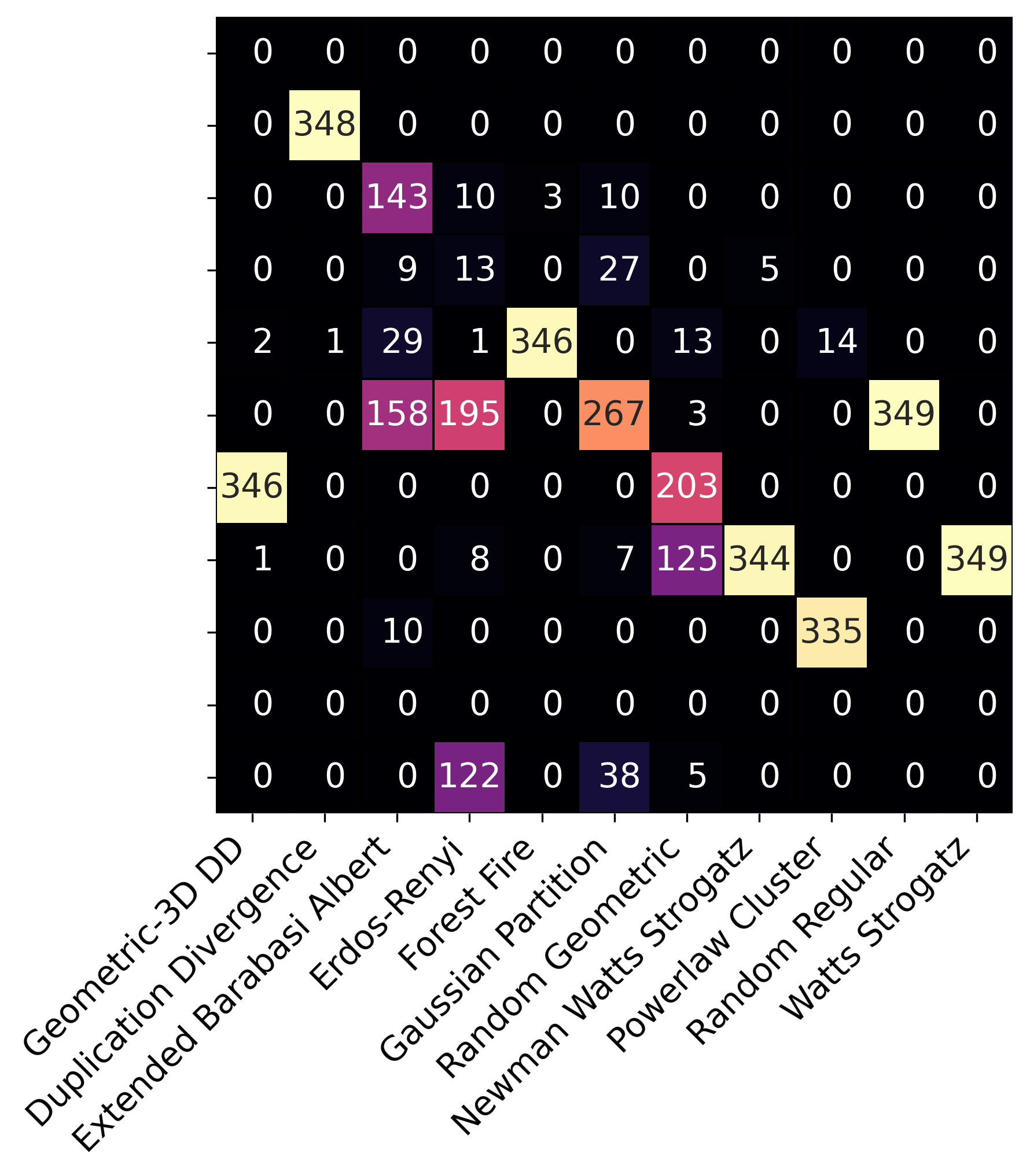}
    \caption{SAGE ND}
    \label{fig:ndsage}
  \end{subfigure}\hspace{0.02\textwidth}%
  \begin{subfigure}[t]{0.23\textwidth}
    \includegraphics[width=1.1\linewidth]{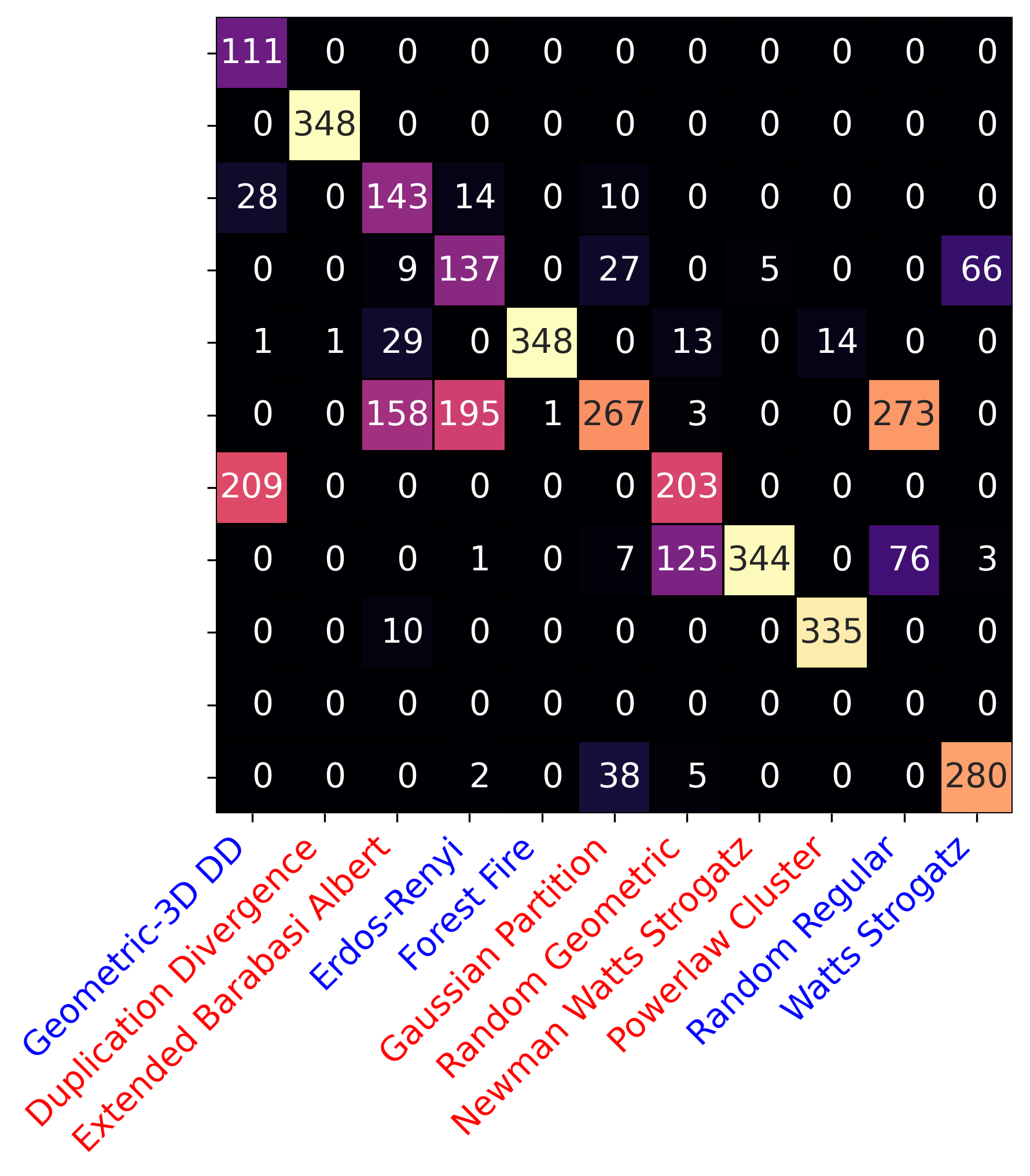}
    \caption{Best ND}
    \label{fig:gin+sage}
  \end{subfigure}
  \caption{Agreement between predictions and true significance-profiles. Last image combines best of \textcolor{red}{SAGE} and \textcolor{blue}{GIN}.}
  \label{fig:heatmaps}
\end{figure}

Figure~\ref{fig:heatmaps} presents confusion-matrix style heatmaps, $H$,  with respect to the graph generator for the agreement between the predicted and true significance-profiles (SPs). Let exist a graph $\gG$, model $M$, graph generators $X$ and $Y$ and an $\sS$, denoting the set of all possible true SPs from the used datasets. Let $M(\gG) = \hat{\vs}$ and $\vs_k = \operatorname{arg min}_{\vs_i \in \sS} d(\hat{\vs}, \vs_i)$ where $d$ is the mean absolute difference.
If the $X$ is the generator of the graph that originates $\vs_k$ and $Y$ the generator from the graph that lead to $\hat{\vs}$, then $H[X,Y] \leftarrow H[X,Y] +1$. Hence, a large number on the diagonal means that the generator did a good job at predicting SPs that can be traced back to the correct graph generator.

Following Figure~\ref{fig:heatmaps}, we conclude that the predictions made by all the models are reasonable for all generators. The results suggest that the model is sufficiently expressive to distinguish between different graph generators, as predictions often align with the correct graph generator. In fact, for Figure~\ref{fig:dgin}, the errors mainly originate from mismatches between generators that are fundamentally very similar. For example, the lollipop graph can be considered a special case of the barbell graph. The same applies for the balanced and full rary tree. As for Figure~\ref{fig:ndgin} and~\ref{fig:ndsage}, the same applies. For example, the Watts-Strogatz graphs are often confused with the Newman-Watts-Strogatz graphs. Furthermore, we note that graphs that are known to completely impossible to be distinguished by 1-WL models like the Random Regular are always misrepresented. 
It is worth noting that SAGE and GIN extract different knowledge from the graphs. In fact, they seem complement each other rather well, as it can be seen in Figure~\ref{fig:gin+sage}. 
Assuming that the choice of null model had little impact in the difficulty of the problem, the conclusion of the ability of a model with expressivity $\leq$ 1-WL to be able to distinguish graphs of different generators (inter-generator predictions) can be seen as a partial empirical confirmation of an old result by Babai and Kucera~\cite{BabaiKucera1979, Babai1980}, regarding the 1-WL test being able to distinguish any random graph with high probability as the size of graph approaches infinity.

Tables~\ref{tbl:corr-vs-incorr-d} and~\ref{tbl:corr-vs-incorr-nd}, give a detailed look at correct predictions as enunciated in Section~\ref{sec:methodology}.
Overall, for this type of analysis, the performance of the models is reliable for some graph families but exhibits systematic errors in others. For example, the powerlaw cluster, forest-fire and duplication-divergence for the ND and star-graphs and circular-ladders for D present good performances. Some allow for $\geq75\%$ correct predictions at $25\%$ error, and all easily beat the baseline $T$. However, generally, most of the predictions do not accurately represent the true SPs.
The conclusion of the inability of the models to generally distinguish graphs with high granularity among those in the same generator (intra-generator predictions) has theoretical backing for the case of the random regular generator~\cite{BabaiKucera1979, Cai1989, Babai1980}. As for the other generators, following the result in~\cite{BabaiKucera1979, Babai1980}, theoretically, it should be highly probable that a model as powerful as the 1-WL could distinguish most of the graphs, not only at inter-generator level, but also at intra-generator level. 
We hypothesise that the mentioned result might not be very useful in practice. However, due to the good results for some generators, it is possible that size of graphs used is not enough for the bad performing generators.

\begin{table}[ht]
\centering
\caption{Number of D graphs with a predicted SP deemed ``correct'' (C) or ``incorrect'' (I).  Generators with \fcolorbox{white}{orange!30}{$\geq50\%$} or \fcolorbox{white}{red!30}{$\geq40\%$} are highlighted.}
\label{tbl:corr-vs-incorr-d}

\renewcommand{\arraystretch}{1}  
\setlength{\tabcolsep}{15pt}  
\resizebox{0.95\textwidth}{!}{  
\begin{tabular}{lcccccccc}
\toprule
                   & \multicolumn{8}{c}{GIN} \\ 
                   & \multicolumn{2}{c}{5\%} & \multicolumn{2}{c}{10\%} & \multicolumn{2}{c}{25\%} & \multicolumn{2}{c}{50\%} \\
                   & C  & I         & C   & I         & C   & I         & C   & I \\
\midrule
Balanced Tree      & 13 & 307       & 14  & 306       & 39  & 281       & 52  & 268 \\
Barbell            & 63 & 257       & 112 & 208       & \cellcolor{red!30}138 & \cellcolor{red!15}182 & \cellcolor{red!30}143 & \cellcolor{red!15}177 \\
Binomial Tree      & 0  & 320       & 0   & 320       & 0   & 320       & 2   & 318 \\
Chordal Cycle      & 0  & 320       & 55  & 265       & 104 & 216       & 114 & 206 \\
Circular Ladder    & 0  & 320       & \cellcolor{red!30}145 & \cellcolor{red!15}175 & \cellcolor{red!30}152 & \cellcolor{red!15}168 & \cellcolor{red!30}152 & \cellcolor{red!15}168 \\
DGM Graph          & 0  & 320       & 67  & 253       & 114 & 206       & 116 & 204 \\
Full Rary Tree     & 4  & 316       & 21  & 299       & 34  & 286       & 36  & 284 \\
Square Lattice     & 0  & 320       & 65  & 255       & 66  & 254       & 66  & 254 \\
Hexagonal Lattice  & 0  & 320       & 0   & 320       & 0   & 320       & 3   & 317 \\
Lollipop           & 0  & 320       & 0   & 320       & 100 & 220       & 100 & 220 \\
Star               & 6  & 152       & \cellcolor{red!30}70  & \cellcolor{red!15}88  & \cellcolor{red!30}76  & \cellcolor{red!15}82  & \cellcolor{orange!30}83  & \cellcolor{orange!15}75 \\
Triangular Lattice & 35 & 285       & 117 & 203       & \cellcolor{red!30}143 & \cellcolor{red!15}177 & \cellcolor{red!30}144 & \cellcolor{red!15}176 \\
\bottomrule
\end{tabular}
}
\end{table}

\begin{table}[ht]
\centering
\caption{Number of ND graphs with a predicted SP deemed ``correct'' (C) or ``incorrect'' (I). Generators with \fcolorbox{white}{green!30}{$\geq90\%$}, \fcolorbox{white}{cyan!30}{$\geq75\%$} or \fcolorbox{white}{orange!30}{$\geq50\%$} are highlighted.}
\label{tbl:corr-vs-incorr-nd}

\renewcommand{\arraystretch}{1.3}  
\setlength{\tabcolsep}{8pt}  

\resizebox{1\textwidth}{!}{  
\begin{tabular}{lcccccccc|cccccccc} 
\toprule
                         & \multicolumn{8}{c|}{GIN}                                                              & \multicolumn{8}{c}{SAGE}                                                             \\ 
                         & \multicolumn{2}{c}{5\%} & \multicolumn{2}{c}{10\%} & \multicolumn{2}{c}{25\%} & \multicolumn{2}{c|}{50\%} & \multicolumn{2}{c}{5\%} & \multicolumn{2}{c}{10\%} & \multicolumn{2}{c}{25\%} & \multicolumn{2}{c}{50\%}  \\
                         & C  & I      & C   & I      & C   & I      & C   & I      & C  & I      & C  & I      & C   & I      & C  & I       \\ 
\midrule
Geometric-3D DD          & 0  & 349    & 67  & 282    & \cellcolor{orange!30}246 & \cellcolor{orange!15}103 & \cellcolor{orange!30}246 & \cellcolor{orange!15}103 & 0  & 349    & 3   & 346    & 45  & 304    & 53  & 296  \\
Duplication Divergence   & 0  & 349    & \cellcolor{orange!30}246 & \cellcolor{orange!15}103 & \cellcolor{cyan!30}293 & \cellcolor{cyan!15}56  & \cellcolor{cyan!30}293 & \cellcolor{cyan!15}56  & 10 & 339    & 33  & 316    & 42  & 307    & 44  & 305  \\
Extended Barabasi Albert & 0  & 349    & 16  & 333    & 62  & 287   & 63  & 286   & 0  & 349    & 41  & 308    & \cellcolor{orange!30}224 & \cellcolor{orange!15}125 & \cellcolor{orange!30}226 & \cellcolor{orange!15}123  \\
Erdos-Renyi              & 0  & 349    & 0   & 349    & 0   & 349   & 9   & 340   & 0  & 349    & 0   & 349    & 0   & 349   & 11  & 338  \\
Forest Fire              & 1  & 348    & 50  & 299    & \cellcolor{orange!30}261 & \cellcolor{orange!15}88  & \cellcolor{green!50}335 & \cellcolor{green!25}14  & 4  & 345    & 100 & 249    & \cellcolor{green!50}323 & \cellcolor{green!25}26  & \cellcolor{green!50}334 & \cellcolor{green!25}15  \\
Gaussian Partition       & 0  & 349    & 0   & 349    & 14  & 335   & 32  & 317   & 0  & 349    & 0   & 349    & 1   & 348   & 15  & 334  \\
Random Geometric         & 0  & 349    & 27  & 322    & 86  & 263   & 86  & 263   & 0  & 349    & 1   & 348    & 34  & 315   & 113 & 236  \\
Newman Watts Strogatz    & 0  & 349    & 126 & 223    & 134 & 215   & 145 & 204   & 0  & 349    & 0   & 349    & 161 & 188   & \cellcolor{orange!30}222 & \cellcolor{orange!15}127  \\
Powerlaw Cluster         & 0  & 349    & \cellcolor{orange!30}176 & \cellcolor{orange!15}173 & \cellcolor{green!50}349 & \cellcolor{green!25}0   & \cellcolor{green!50}349 & \cellcolor{green!25}0   & 97 & 252    & \cellcolor{cyan!30}301 & \cellcolor{cyan!15}48  & \cellcolor{green!50}349 & \cellcolor{green!25}0   & \cellcolor{green!50}349 & \cellcolor{green!25}0   \\
Random Regular           & 0  & 349    & 0   & 349    & 0   & 349   & 0   & 349   & 0  & 349    & 0   & 349    & 0   & 349   & 61  & 288  \\
Watts Strogatz           & 0  & 349    & 29  & 320    & 74  & 275   & 97  & 252   & 0  & 349    & 0   & 349    & 43  & 306   & 56  & 293  \\
\bottomrule
\end{tabular}%
}
\end{table}

\subsection{Validations of the Assumptions Made}

We begin by validating whether the assumption that predicting multiple scores simultaneously yields benefits over predicting each score individually.
To do this, we trained eight models, each corresponding to a graph in $\Omega$, using the ND segment. We employed the GIN variant of MPNNs, chosen for its theoretical and practical advantages. The models were trained without prior assumptions so that, when training for a single prediction, each model specialized in its respective graph. Table~\ref{tbl:single-vs-multi} presents the percentiles of the squared difference between true and predicted SPs on the validation dataset for each graph.

\begin{table}[ht]
\centering
\caption{Error percentiles in the validation set of the squared error and their percent \textcolor[rgb]{0,0.502,0}{decrease}/\textcolor{red}{increase} comparing the multitarget to the single target model.}
\label{tbl:single-vs-multi}

\renewcommand{\arraystretch}{1.1}  
\setlength{\tabcolsep}{8pt}  

\resizebox{1\textwidth}{!}{  
\begin{tabular}{ccccccccc} 
\toprule
\textbf{Graph}         & \textbf{\graphSeven}~                           & \textbf{\graphSix}                             & \textbf{\graphFive}                                 & \textbf{\graphFour}                                                        & \textbf{\graphThree}                             & \textbf{\graphTwo}                           & \textbf{\graphOne}                             & \textbf{\graphZero}                             \\ 
Type                   & single~ multi                         & single~ multi                          & single~ multi                              & single~ multi                                                     & single~ multi                          & single~ multi                        & single~ multi                          & single~ multi                          \\ 
\midrule
\multirow{2}{*}{100\%} & 2.921~ 2.868                          & 1.231~ 0.991                           & 1.193~ 0.816                               & 1.498~ 0.861                                                      & 1.056~ 1.268                           & 2.275~ 2.388                         & 1.814~ 2.193                           & 2.940~ 2.621                           \\
                       & \textcolor[rgb]{0,0.502,0}{-1.814\%}  & \textcolor[rgb]{0,0.502,0}{-19.496\%}  & \textcolor[rgb]{0,0.502,0}{-31.601\%}      & \textcolor[rgb]{0,0.502,0}{-42.523}\textcolor[rgb]{0,0.502,0}{\%} & \textcolor{red}{+20.076\%}             & \textcolor{red}{+4.967\%}            & \textcolor{red}{+20.893\%}             & \textcolor[rgb]{0,0.502,0}{-10.850\%}  \\
\multirow{2}{*}{95\%}  & 0.279~ 0.250                          & 0.316~ 0.396                           & 0.309~ 0.320                               & 0.350~ 0.347                                                      & 0.547~ 0.502                           & 0.517~ 0.489                         & 1.294~ 0.741                           & 1.463~ 0.808                           \\
                       & \textcolor[rgb]{0,0.502,0}{-10.394\%} & \textcolor{red}{+25.316\%}             & \textcolor{red}{+3.560\%}                  & \textcolor[rgb]{0,0.502,0}{-0.857\%}                              & \textcolor[rgb]{0,0.502,0}{-8.227\%}   & \textcolor[rgb]{0,0.502,0}{-5.416\%} & \textcolor[rgb]{0,0.502,0}{-42.736\%}  & \textcolor[rgb]{0,0.502,0}{-44.771\%}  \\
\multirow{2}{*}{75\%}  & 0.078~ 0.091                          & 0.095~ 0.043                           & 0.047~ 0.041                               & 0.083~ 0.053                                                      & 0.067~ 0.038                           & 0.082~ 0.100                         & 0.210~ 0.042                           & 0.357~ 0.048                           \\
                       & \textcolor{red}{+16.667\%}            & \textcolor[rgb]{0,0.502,0}{-54.737\%}  & \textcolor[rgb]{0,0.502,0}{-12.766\%}      & \textcolor[rgb]{0,0.502,0}{-36.145\%}                             & \textcolor[rgb]{0,0.502,0}{-43.284\%}  & \textcolor{red}{+21.951\%}           & \textcolor[rgb]{0,0.502,0}{-80.000\%}  & \textcolor[rgb]{0,0.502,0}{-86.555\%}  \\
\multirow{2}{*}{50\%}  & 0.004~ 0.009                          & 0.017~ 0.011                           & 0.004~ 0.006                               & 0.008~ 0.007                                                      & 0.005~ 0.003                           & 0.007~ 0.012                         & 0.051~ 0.007                           & 0.042~ 0.007                           \\
                       & \textcolor{red}{+125.000\%}           & \textcolor[rgb]{0,0.502,0}{-35.294\%}  & \textcolor{red}{+50.000\%}                 & \textcolor[rgb]{0,0.502,0}{-12.500\%}                             & \textcolor[rgb]{0,0.502,0}{-40.000\%}  & \textcolor{red}{+71.429\%}           & \textcolor[rgb]{0,0.502,0}{-86.275\%}  & \textcolor[rgb]{0,0.502,0}{-83.333\%}  \\
\multirow{2}{*}{25\%}  & 0.000~ 0.001                          & \multicolumn{1}{l}{0.001~ 0.000}       & 0.000~ 0.000                               & 0.001~ 0.001                                                      & 0.001~ 0.000                           & 0.001~ 0.002                         & 0.005~ 0.000                           & 0.012~ 0.003                           \\
                       & \textcolor{red}{+100.000\%}           & \textcolor[rgb]{0,0.502,0}{-100.000\%} & \textcolor[rgb]{0.502,0.502,0.502}{0.00\%} & \textcolor[rgb]{0.502,0.502,0.502}{0.00\%}                        & \textcolor[rgb]{0,0.502,0}{-100.000\%} & \textcolor{red}{+100.000\%}          & \textcolor[rgb]{0,0.502,0}{-100.000\%} & \textcolor[rgb]{0,0.502,0}{-75.000\%}  \\
\bottomrule
\end{tabular}
} %
\end{table}

Based on Table~\ref{tbl:single-vs-multi}, multi-target regression generally improves prediction accuracy across all graphs except for, arguably, the four-node clique and four-path. This supports our expectation, outlined in Section~\ref{sec:method}, that jointly predicting graphs with shared traits enhances performance.
Regarding the increased error observed in the other two graphs, we hypothesise that this may be due to their limited benefit from shared information, as other graphs lack sufficient encoded data to enhance their predictions beyond a specialized model. Nevertheless, given the relative magnitude of error changes, we conclude that multi-target regression is superior for motif estimation. Additionally, it offers the advantage of requiring only a single model while maintaining competitive training times compared to single-target regression.

The second assumption we must validate is whether estimating SP directly is beneficial over calculating the SP after estimating subgraph counts.
To do this, we trained a model, using the ND segment, to predict the frequency of graphs in $\Omega$ directly. Model and training procedure remain as in the former comparison.

To avoid generating 500 random networks per test instance, we opted for approximating the Z-score that would be later obtained from subgraph estimation. To achieve this, we decomposed the subgraph frequency variable into actual frequency ($\ry$) and model error ($\rz$).
Assuming that the difference between a value $z \sim \rz$ and $\mu_\rz$ is proportional to $\sigma_\rz$, minding signal indetermination, following $(y-\mathbb{E}[\ry])\pm\sigma_\rz \, / \,\big(Var(\ry)^2+Var(\rz)^2\big)^{1/2}$, we do not have to do additional calculations since all values were either acquired during the training of the model (values regarding $\rz$) or were collected during the dataset construction (values regarding $\ry$).
Table~\ref{tbl:count-vs-sp} presents percentiles for the absolute difference between true and predicted SPs. The values under ``Count'' correspond to the minimum difference (hence, worst case comparison) resulting from estimations with all valid signal combinations.

\begin{table}[ht]
\centering
\caption{Error percentiles in the real-world dataset of the absolute difference and their percent \textcolor[rgb]{0,0.502,0}{decrease}/\textcolor{red}{increase} comparing direct SP estimation (SP) to estimating graph frequencies (Count).}
\label{tbl:count-vs-sp}

\renewcommand{\arraystretch}{1.1}  
\setlength{\tabcolsep}{8pt}  

\resizebox{1\textwidth}{!}{  
\begin{tabular}{ccccccccc} 
\toprule
\textbf{Graph}        & \textbf{\graphSeven}                            & \textbf{\graphSix}                  & \textbf{\graphFive}                            & \textbf{\graphFour}                 & \textbf{\graphThree}                  & \textbf{\graphTwo}                            & \textbf{\graphOne}                            & \textbf{\graphZero}                             \\ 
Type                  & Count~ SP                             & Count~ SP                   & Count~ SP                             & Count~ SP                  & Count~ SP                   & Count~ SP                             & Count~ SP                             & Count~ SP                              \\ 
\cmidrule{2-9}
\multirow{2}{*}{75\%} & 0.469 0.222  & 0.377~ 0.323 & 0.326~ 0.199 & 0.322~ 0.227 & 0.361 0.173 & 0.680~ 0.298 & 0.509~ 0.172 & 0.625~ 0.278                           \\
                      & \textcolor[rgb]{0,0.502,0}{-52.584\%} & \textcolor[rgb]{0,0.502,0}{-14.378\%}  & \textcolor[rgb]{0,0.502,0}{-38.799\%}            & \textcolor[rgb]{0,0.502,0}{-29.618\%} & \textcolor[rgb]{0,0.502,0}{-52.117\%} & \textcolor[rgb]{0,0.502,0}{-56.225\%} & \textcolor[rgb]{0,0.502,0}{-66.154\%} & \textcolor[rgb]{0,0.502,0}{-55.554\%}  \\

\multirow{2}{*}{50\%} & 0.339~ 0.116 & 0.236 0.109  & 0.175~ 0.056 & 0.174~ 0.083 & 0.214 0.083 & 0.311~ 0.126 & 0.368~ 0.072 & 0.358~ 0.079                           \\
                      & \textcolor[rgb]{0,0.502,0}{-65.907\%} & \textcolor[rgb]{0,0.502,0}{-53.718\%} & \textcolor[rgb]{0,0.502,0}{-68.007\%} & \textcolor[rgb]{0,0.502,0}{-52.079\%} & \textcolor[rgb]{0,0.502,0}{-61.143\%}  & \textcolor[rgb]{0,0.502,0}{-59.456\%} & \textcolor[rgb]{0,0.502,0}{-80.448\%} & \textcolor[rgb]{0,0.502,0}{-78.013\%}  \\

\multirow{2}{*}{25\%} & 0.041~ 0.044 & 0.115~ 0.042 & 0.106~ 0.036 & 0.065~ 0.031 & 0.063 0.020 & 0.144~ 0.041 & 0.328~ 0.027 & 0.300~ 0.021                          \\
                      & \textcolor{red}{+5.426
\%} & \textcolor[rgb]{0,0.502,0}{-63.960\%}  & \textcolor[rgb]{0,0.502,0}{-66.396\%}           & \textcolor[rgb]{0,0.502,0}{-52.797\%} & \textcolor[rgb]{0,0.502,0}{-68.212\%}   & \textcolor[rgb]{0,0.502,0}{-71.614\%} & \textcolor[rgb]{0,0.502,0}{-91.903\%} & \textcolor[rgb]{0,0.502,0}{-92.887\%}  \\
\bottomrule
\end{tabular}
} %
\end{table}

Results show a significant improvement when using direct estimation for all graphs. Given its gain and computational efficiency (not having to directly calculate the occurrences for the control graphs), direct significance-profile prediction is preferable for motif estimation in the chosen null model.



\subsection{Model Predictions in the Real-World Dataset}\label{sub:model-predictions-real}

Similarly to the synthetic data, the predictions on the real-world dataset are, generally, not very good at precise intra-category level estimation. Interestingly, models grouped real-world graphs based on their ``similarity'' to synthetic datasets. For example, consider the \textit{ia-escorts-dynamic} network (Figure~\ref{fig:mlreal-inter-nd-gin}). In this network, nodes represent buyers and escorts, and edges indicate interactions between them. Its SP profile closely resembled that of a duplication-divergence model.
Next, take the \textit{coauthor-CS} network (Figure~\ref{fig:mlreal-colabcit-nd-gin}). Here, nodes represent authors, and an edge connects two nodes if the authors have co-authored a paper. This network produced an SP profile similar to that of a forest-fire model.
Finally, consider the \textit{ia-primary-school-proximity} network (Figure~\ref{fig:mlreal-inter-nd-gin}). In this case, nodes represent individuals, and an edge is formed when two people are in close physical proximity for a certain period. The SP profile for this network matched that of a geometric model.
Among other correct matches, these three examples demonstrate a concrete case where the model found the appropriate synthetic generator for the real-world network\footnote{More details about these networks can be found in the supplemental material} based on the significance-profile. 
This suggests that while models struggle with precise real-world predictions, they can help identify the closest synthetic model for real networks based on significance-profiles. (Images of predictions available in the additional material). Additionally, the model remains stable, with errors increasing by at most $\approx20\%$ as networks scales up to 1000 times the train size.

\textbf{Points of divergence.} In network similarity discovery based on SP, two key concepts are crucial. First, the model’s ability to distinguish networks is constrained by the expressivity of the space of the SP used. The more expressive the space, the more reliable the ability of the model to distinguish networks. Secondly, if the model predicts similar profiles for two graphs $\gG$ and $\gH$, indicating they resemble a graph $\gF$, this suggests $\gG$ and $\gH$ may originate from a process similar to $\gF$. However, this conclusion is valid only if the true profiles of $\gG$ and $\gH$ are indeed similar; otherwise, the model’s lack of expressivity leads to incorrect conclusions.


\subsection{Time Comparisons}\label{sub:time}

We will compare the efficiency of doing predictions using a GNN model with using Gtrie. We will employ the normalized measures of Speedup, $\text{Speedup} = \text{Total Time Used}^{(A)}/\text{Total Time Used}^{(B)}$ and Core Efficiency Gain $\text{Core Efficiency Gain} = \text{Total Core Time}^{(A)}/\text{Total Core Time}^{(B)}$ for tasks $A$ and $B$.

The GNN model presented a $\sim 1.6$ million speedup for the medium-large real dataset with a $\sim 1625$ core efficiency gain. For the small real dataset, the values were $\sim 18211$ speedup and a $\sim 19$ core efficiency gain. For the deterministic segment we got a $\sim 82594$ speedup and a $\sim 112$ times more efficiency per core. As for the non-deterministic segment, we got a $\sim 548888$ speedup and a $\sim 745$ more core efficient task.

\section{Conclusions}\label{sub:chap4-conc}

Although no GNN-based method is specifically designed for predicting motifs, our MPNN models with synthetic data still fall short for precise real-world SP discovery.
However, we empirically showed that through simple modifications, namely, multitarget regression and direct SP estimation, we achieved stability across out-of-distribution network sizes and improved results over existing traditional SP estimation via subgraph counting. Additionally, we hint that when performing direct SP estimation, the incorporation of a null model may help overcome the theoretical limitations of motif estimation based on subgraph counting. Our findings also suggest that the models are promising for network categorisation since they can distinguishing high-level differences between graphs. Future work can focus on categorising the SP estimation problem space based on different null models.



\clearpage
\appendix

\section{Data Details}\label{sec:data-details}

In this section we discuss with more detail some aspects of the data, both synthetic and real-world, used in the experiments discussed in the main text. Section~\ref{sub:nd-gen} presents details of how the non-deterministic synthetic data was generated and Section~\ref{sub:d-gen} details about the deterministic segment. Section~\ref{sub:pattern-connect} details findings on how the significance-profiles behave with respect to each other. Finally, Section~\ref{sub:real-data} makes a brief analysis of the real-world data used.

\subsection{Non-Deterministic Generators}\label{sub:nd-gen}

For the Erd\H{o}s-Renyi model~\cite{Erdos1960a}, we mainly aim at creating graphs in the three (out of four) main topological phases a graph achieve~\cite{Barabasi2017}. 
We exclusively uniformly control the number of nodes within each of the delineated phases, namely, the ``critical'', ``supercritical'' and ``connected'' states. This strategic regulation facilitates substantial variability in graph size while preventing an excessive escalation of the referred metric that could possibly impede further computational processing.  

For the Watts-Strogatz~\cite{Watts1998} and Newman Watts-Strogatz~\cite{Newman1999}, we regulated the generation based on the total number of nodes, the initial number of neighbours and the probability of rewiring in order to generate networks that represented key sections of the characterisation based on the clustering coefficient and path length, as given by Watts and Strogatz~\cite{Watts1998}.

For the extended Barabási-Albert model~\cite{Albert2000a}, we defined as hyperparameters the total number of nodes and amount of connections a new node gains. Based on the equations delineated in the original article, we modulate our parameters grounded on two variables: the probability associated with the formation of new links ($p$) and the probability of rewiring existing connections ($q$). We aim to have graphs characterised by a power-law degree distribution with an exponent ranging uniformly between $2$ and $3$.

For the cluster power-law~\cite{Holme2002}, we vary uniformly the number of nodes and calculate the necessary probability according to the original study to obtain a clustering coefficient of $0.35$, $0.45$ or $0.55$.

The duplication-divergence generator~\cite{Ispolatov2005} operates by randomly selecting a node $v$ from an initial graph and duplicating all edges connected to $v$ with a retention probability denoted as $\sigma$. We select three regimes. Two of the selected regimes exhibit self-averaging behaviour concerning the number of edges, specifically when $0 < \sigma < e^{-1}$ and $e^{-1} < \sigma < 1/2$. The non-self-averaging regime is characterised by $1/2 < \sigma < 1$. 

In the Gaussian random partition model~\cite{Brandes2003-GRP}, $k$ groups of nodes are generated with $t$ nodes derived from a Gaussian distribution with mean $s$ and variance $v$. The connectivity between nodes in a group is given by a probability $p$, and the connectivity inter-groups is given by $q$. In this generator, we parameterise the number of nodes $|V|$, the size of the $k$ groups and the maximum number of allowed edges $|E_{\text{max}}|$. Both the $p$ and $q$ probabilities are calculated to not exceed the maximum number of edges according to~\Eqref{eq:grp}.

\begin{align}\label{eq:grp}
    q &\leq min\Bigg(1, \; \frac{2|E_{\text{max}}|}{|V|^2 + |V|\big(\kappa \cdot s^{1/2} - s(\kappa + 1)\big)} \Bigg) \\
    p &\leq  min(1, \; \kappa \cdot q)
\end{align}

We defined $p$ as always having the possibility of going above $q$ because we would like to have networks that can have a community structure in order to have a more diverse set of graphs. Hence, we put the bound of $p$ as being scaled over $q$ by $\kappa$, which we called over-attractiveness. The values used for the $v$ and $\kappa$ are $10$ and $5$ respectively. All other parameters are uniformly sampled from a predefined range\footnote{More details for the parameters available in the code.}.

In the case of the forest-fire model~\cite{Leskovec2007}, we varied the number of nodes and the backward and forward probability between $0$ and $0.4$ (inclusive) to try to steer away from very aggressive Densification Power Law exponents and clique-like graphs, characteristics that, if severe, can hinder the subsequent steps from a computational point of view. With the values for the probabilities defined above, we expect to observe sparse networks that slowly ``densify over time'', together with decreasing diameter. All the graphs are made undirected after being generated.

For the random geometric graph, since some properties of the graph related to its connectedness, such as maximum cluster size and coefficient, vary with the dimension of the unit hypercube used~\cite{Dall2002, Penrose2003-du}, besides the number of nodes, we we decided to vary the dimension of the hypercube between $2$ and $5$. However, we did not efficiently explore all possible configurations within the referred dimensions because we limited the number of edges. Similarly to a random geometric model, we used a random geometric model in 3D with duplication divergence~\cite{Higham2008-3DDD}. For this model, we followed Silva et al.~\cite{Silva2023-DG}.

The last model in the non-deterministic segment is the random regular generator.
In this case, the parameters subject to uniform variation were the total number of nodes and the degree assigned to each node, which once determined, remain constant across all nodes.

\subsection{Deterministic Generators}\label{sub:d-gen}

We complemented the graphs generated by the non-deterministic generators with smaller amounts of graphs from deterministic generators. These generators have their network completely and without randomness determined once their parameters are chosen. 

The first group of deterministic generators consists of multiple types of trees. 
We use the binomial tree model parametrised on its order and the balanced tree (full rary-tree) parametrised on its height and branching factor.

The second group is based on modified cycles. We use the circular ladder generator, varying the complete size of the graph and the chordal cycle~\cite{Lubotzky1994-CHORDAL}, also varying its complete size.

The third group is based on complete graphs and encompasses the barbell and lollipop graphs. The barbell graph is made of two complete graphs of size $k$ connected by a path of size $m$. The lollipop is a barbell graph with only one complete graph and the path. In order to not complicate subsequent steps, we carefully limited the size of the complete graphs.

The fourth group consists of the Dorogovtsev-Goltsev-Mendes model~\cite{Dorogovtsev2002-model}. This generator modulates a scale-free discrete degree distribution with exponent $1+ ln \, 3/ ln \, 2$ by using a rather simple rule: ``\textit{At each time step, to every edge of the graph, a new vertex is added, which is attached to both the end vertices of the edge.}'' (in~\cite{Dorogovtsev2002-model}). We vary the magnitude of the number of nodes and edges by changing $n$, resulting in $3(3^{n} + 1) / 2$ and $3^{n+1}$ nodes and edges respectively.

The fifth group consists of lattices. Namely, we use 2D hexagonal, triangular lattices and 3D square lattices. The first $2$ lattices have the option of allowing for boundary periodicity. All lattices vary in terms of the size of each dimension.

Finally, the last group consists of star graphs of various sizes.

Since the types of graphs that the deterministic generators generate are not subject to randomness, it is redundant to create multiple graphs for each set of parameters. However, in order to introduce a degree of randomness to the deterministic graphs, we introduced a probability of random rewiring of a percentage of edges after the graph is generated. The rewiring procedure for a single edge consists of selecting an edge $(u,v)$ from a graph $\gG$, deleting it and attaching one of the ends, $u$ or $v$, to another node $w$. If $u$ is picked and $(u,w)$ already exists, then $\gG$ will exit the procedure with one less edge. Since we want some variability but still want to preserve the original deterministic graphs, for each generator, two sets of graphs $\sS_1$ and $\sS_2$ will be constructed according to the proposed generator parameters mentioned above. After that, $\sS_1$ is not subject to any rewiring, and for each graph in $\sS_2$, $p$\% of its edges are rewired according to the procedure described earlier. According to this methodology, we generated four versions. The first had 2 edges swapped, the second $25\%$ of the edges swapped, the third $10\%$ and the fourth $60\%$. We stick to version two due to being the best performing one according to preliminary tests. This fact means that $25\%$ seems to be a good choice of random-rewiring so that the information encoded in the deterministic graphs is maximised.

\subsection{Pattern interconnectivity}\label{sub:pattern-connect} 



In the Section~\ref{sec:data} of the main text we introduced a result regarding the symmetry of the score of size three patterns. We have a strong indication that even without the normalisation of the Z-score, the number of occurrences between connected graphs of the same size is highly related. More formally, the relation between the Z-scores of the graphs of size three can be described as follows. Let $\rx$ be a random variable denoting the number of induced occurrences of triangles in any graph that follows the degree distribution $D$. Let $\ry$ be a random variable denoting the occurrence of induced 3-paths in any graph that follows the degree distribution $D$. \Eqref{eq:triang-path-rel} gives the relation between Z-scores of $\rx$ and $\ry$.

\begin{align}\label{eq:triang-path-rel}
X &= \begin{cases}
        0, \qquad &\text{if} \quad \ry-\mu_\ry = 0     \\
        -\frac{\sigma_\rx}{\sigma_\ry}(Y - \mu_\ry) + \mu_\rx, \qquad & \text{otherwise}
    \end{cases}
\end{align}

When standardised to a mean of $0$ and a standard deviation of $1$, the Z-scores of both variables, exhibit symmetry. Hence, it is possible to express their non-standardised values as linear combinations of each other. Considering the mean and standard deviation of the counts of 3-paths and triangles for $D$, given any graph $\gG$ that follows $D$, it is possible to get the concrete count of triangles from the count of 3-paths and vice-versa. 

Even though from a practical point of view, the result from~\Eqref{eq:triang-path-rel} has little implications due to the dependence on the first raw moment and the second central moment of both the distributions of $\rx$ and $\ry$, it presents a strong indication of what was postulated in Section~\ref{sec:method} of the main text. That is, it further solidifies the connectivity between the graphs selected for $\Omega$. In this case, the relation is so strong that we believe to be redundant to try to predict both scores. Moreover, following the normalisation procedure, the restrained nature of the result raises questions about the choice of modelling the problem as a regression task for the size three graphs. However, despite these observations, we stick to our initial formulation since in theory it does not significantly undermine the capacity of the model.

The result experimentally verified in the above paragraphs can be seen as a small extension of Ginoza and Mugler~\cite{Ginoza2010} and Wegner~\cite{Wegner2014} to undirected patterns of size three. 
In particular, adapting from Wegner,~\Eqref{eq:conservation_equations_3undir} displays the conservation law for the number of induced 3-paths. 

\begin{equation}\label{eq:conservation_equations_3undir}
   \# \text{3-paths}_{ind} = \underbrace{\; \#\text{3-paths}_{\overline{ind}} \;}_{\text{fully defined by degree sequence}} \; - \; \underbrace{\; 3\#\text{triangles} \;}_{\text{not fully defined by degree sequence}}
\end{equation} 

Since the number of non-induced 3-paths depends only on the degree sequence \big($\sum_{i=0}^{|V|} \binom{|N(i)|}{2}$\big), it will not change under the configuration model. Hence, the number of induced 3-paths is a variable that once the degree sequence is fixed, depends only on the number of triangles. As for the number of triangles, they depend on the order the edges are added to the graph under~\Eqref{eq:evolution_equations_3undir}.

\begin{subequations}\label{eq:evolution_equations_3undir}
    \begin{align}
     \text{total triangles} &= \sum_{t=0}^{|E|} \#\text{triangles}_t \\
     \text{total 3-path} &= \sum_{t=0}^{|E|} (\#\text{3-path}_t - \#\text{triangles}_t) \\
     \#\text{triangles}_{t+1} &= |\{w | w \in N(u^t) \land w \in N(v^t) \}| \\
     \#\text{3-path}_{t+1} &= |N(u^t)| + |N(v^t)| - 2\#\text{triangles}_{t+1} 
    \end{align}
\end{subequations}

where nodes $u$ and $v$ represent the nodes that were connected by an edge at iteration $t$. Hence, any realisation of a degree sequence through the configuration model will always have its number of induced 3-paths negatively correlated with the number of triangles. 

\begin{figure}[ht]
    \centering
    \includegraphics[width=1\textwidth]{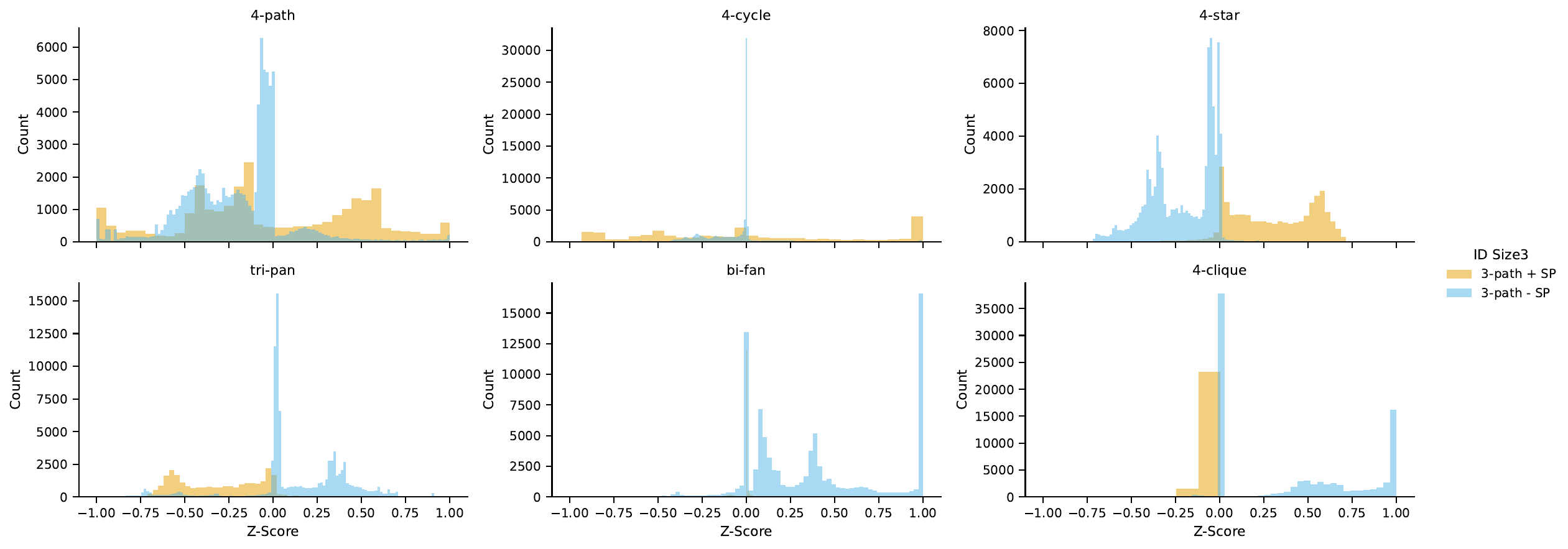}
    \caption{Distribution of the significance profiles for the graphs of size 4, given the value the 3-path took. The positive value corresponds to $1/\sqrt{2}$ and the negative to $-1/\sqrt{2}$.}
    \label{fig:three-influences-four}
\end{figure}

Regarding the relation between graphs of size three and graphs of size four, by analysing Figure~\ref{fig:three-influences-four}, it is possible to understand that there is a relation between the significance-profiles of these graphs. 
This relationship is particularly pronounced concerning the 4-star, tri-pan and 4-clique, as the values of the significance-profiles assumed by these graphs are distributed across the spectrum centred at $0$, contingent upon the value held by the 3-path. For example, if the significance-profile of the 3-path is positive, the significance-profile of the 4-star will most likely be positive following the distribution given by the yellow histogram. As for the 4-cycle and bi-fan, this relation is not as strong. For the 4-cycle, we learn that the values are mostly zero when the significance-profile for the 3-path is negative and is quite dispersed across the space otherwise, with a small peak close to the $1$ value. As for the bi-fan, besides learning that it is unlikely that it takes a large negative value, even though hard to discern from the figure, $46.2\%$ of the values are $0$ when the significance-profile for the 3-path is positive, and $69.7\%$ are between $-0.1$ and $0.1$ for the same conditions.

\subsection{Real-World Dataset}\label{sub:real-data}

The selected type categories are described in list~\ref{lst:real-categories}. The numbers between the square brackets in each bullet point correspond to the number of networks each category has in each of the scale categories (small and medium-large).

\begin{itemize}\label{lst:real-categories}
    \item \textbf{ANIMAL SOCIAL}: [10/8] Networks describing the social behaviour of non-human animals, spanning species such as ants, dolphins, lizards, sheep, and others.
    \item \textbf{BIOLOGICAL}: [10/10]
    Networks of protein-protein interactions, a metabolic networks of small organisms, and a networks of disease connections in humans based on shared genes.
    \item \textbf{BRAIN}: [9/10] 
    Connectome networks of various brain regions such as the cerebral cortex, interareal cortical areas, and synaptic networks, across multiple species including cats, worms, mice, macaques, and humans.
    \item \textbf{CHEMOINFORMATICS}: [10/0] Networks of multiple different enzyme structures.
    \item \textbf{COLLABORATION CITATION} [6/8]: Networks of of paper citations and author collaborations.
    \item \textbf{INFRASTRUCTURE}: [5/7] Electric grids and road networks.
    \item \textbf{INTERACTION}: [5/6] Networks of physical contact between humans in various contexts, together with some digital contact, for example, by e-mail or a phone call.
    \item \textbf{SOCIAL COMMUNICATION}: [2/10] Interaction between humans in social networks such as mutually liked Facebook pages, friendship connections and retweets.
\end{itemize}

\begin{figure}[ht]
    \begin{subfigure}[b]{1\textwidth}
        \centering
        \includegraphics[width=.5\linewidth, align=c]{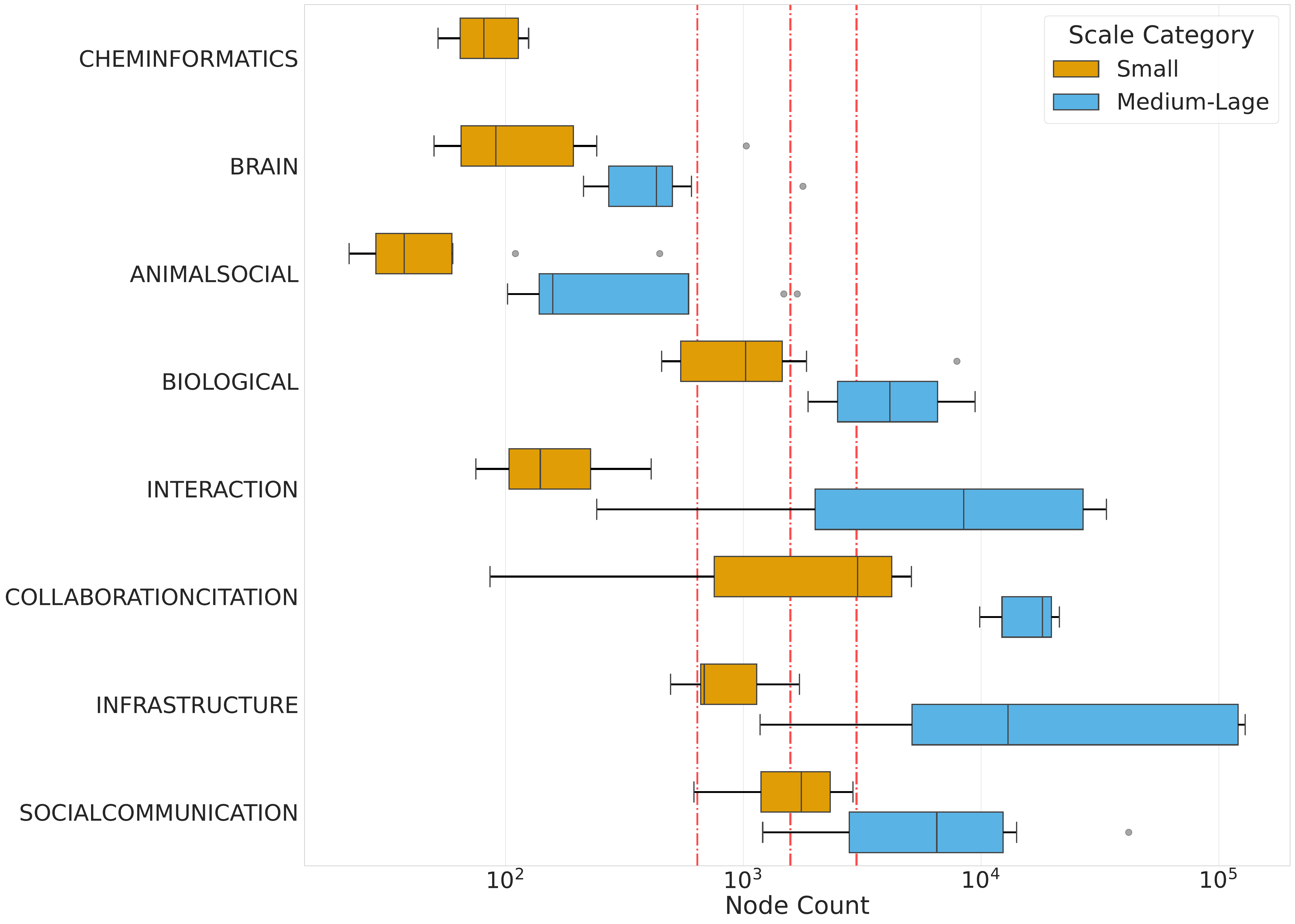}
        \caption{Summary of how the number of nodes is distributed for the real networks.}
        \label{fig:node-real} 
    \end{subfigure}
  \vspace{0.01\textwidth}
    \begin{subfigure}[b]{1\textwidth}
        \centering
        \includegraphics[width=.5\linewidth, align=c]{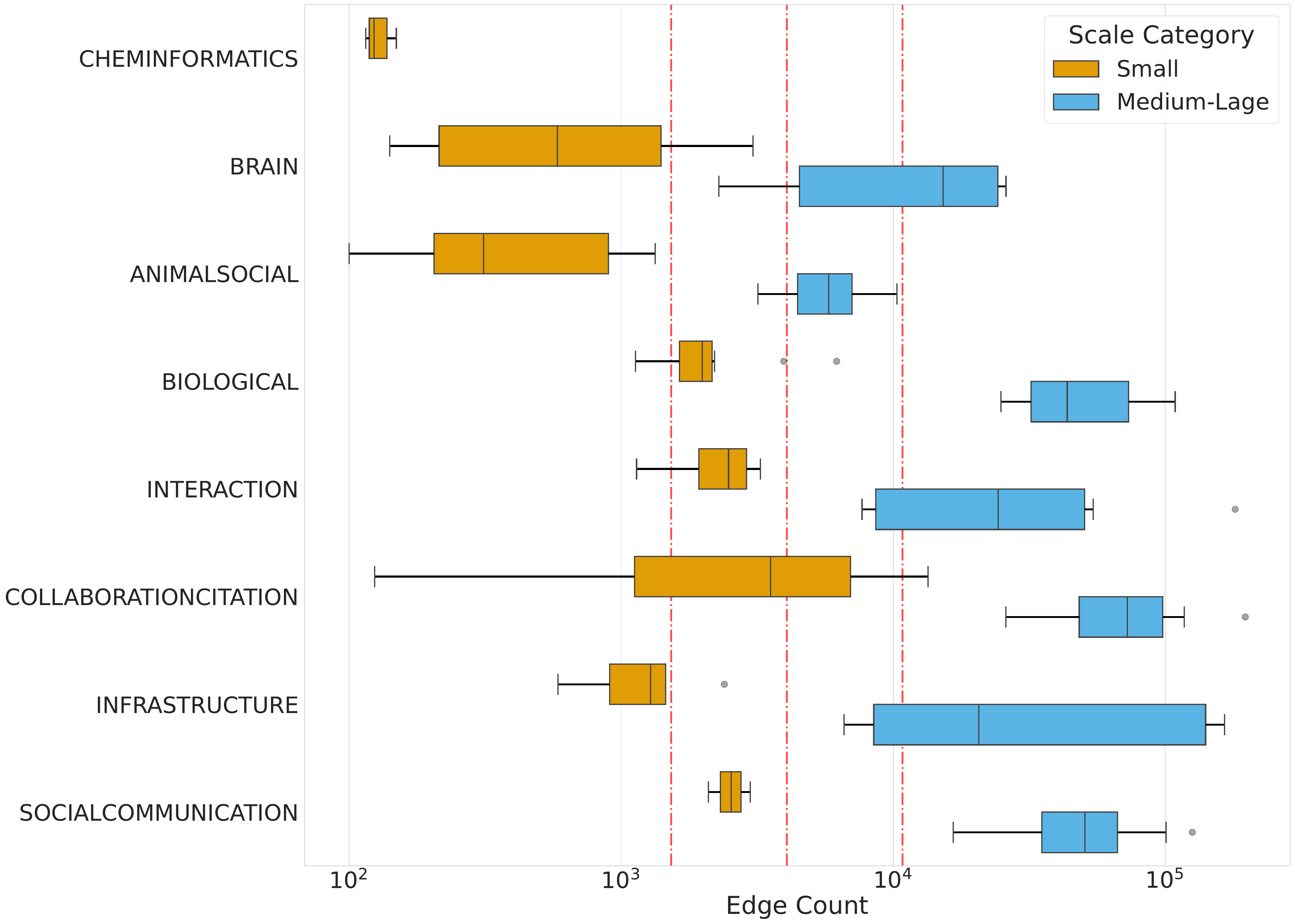}
        \caption{Summary of how the number of edges is distributed for the real networks.}
        \label{fig:edge-real}
    \end{subfigure}
\caption{Summary of the distribution of the node and edge count of the real networks. All data is presented in logarithmic scale.}
\label{fig:barplots-real-data}
\end{figure}

Figure~\ref{fig:node-real} and Figure~\ref{fig:edge-real} show a summary of the number of edges and nodes for the different type and scale categories. The red dashed lines represent the average of the minimum node (or edges) quantity, the average of the mean node (or edges) quantity and the average of the maximum node (or edges) quantity respectively, calculated for all 23 graph models used in the synthetic dataset. 

In Figure~\ref{fig:node-real} and Figure~\ref{fig:edge-real}, we can observe that the distinction between scale categories is influenced by the type category of the network. For example, ANIMAL SOCIAL networks tend to be smaller than INFRASTRUCTURE networks in the medium-large category. However, this relationship varies by scale, as both network types exhibit similar sizes in the small-scale category.
In any case, in the general scenario, we expect that the small-scale category encloses networks from being smaller than an average network from the training set until networks that can be slightly larger than the mean training network. As for the medium-large category, they hold networks that can be around the average size of a training network to networks that are several orders of magnitude larger than the average size of a training network. The dashed red lines in Figure~\ref{fig:node-real} and Figure~\ref{fig:edge-real} help validate this statement. 

In summary, we have $56$ networks in the small-scale category and $59$ in the medium-large category. If a graph in the test set was not already a simple undirected static graph when it was obtained, we transformed it in a graph following said conditions\footnote{Please see the supplemental material for the references for each of the individual 115 networks.}.

\section{Methodological Details}\label{sec:methdology-details}

Section~\ref{sub:data-part} describes how the data was partitioned for training and Section~\ref{sub:base-model} the base model, the hyperparameter space and the method and software used for training.

\subsection{Exact dataset partition}\label{sub:data-part}

We conduct separate experiments using the two segments of the produced data, the deterministic and the non-deterministic. Furthermore, the split in train-validation-test is stratified by random sampling a percentage $p$ of each of the generators for each segment. 
In the case of the deterministic segment, we use all $3200$ graphs available. With $p=0.7$, the training set has $3200\times0.7\times12 = 26880$ graphs, and the validation set has $3200\times0.2\times12=7680$ graphs. The remaining $10\%$ are used for the test set. We avoided using larger datasets due to memory restrictions. As for the non-deterministic segment, in order for it to have a comparable total size to the deterministic segment, we sampled $3490\times0.7\times11=26873$ graphs and the validation set $3490\times0.2\times11=7678$. 

\subsection{Initial Base Model}\label{sub:base-model}

\begin{figure}[ht]
    \centering
    \includegraphics[width=0.85\textwidth]{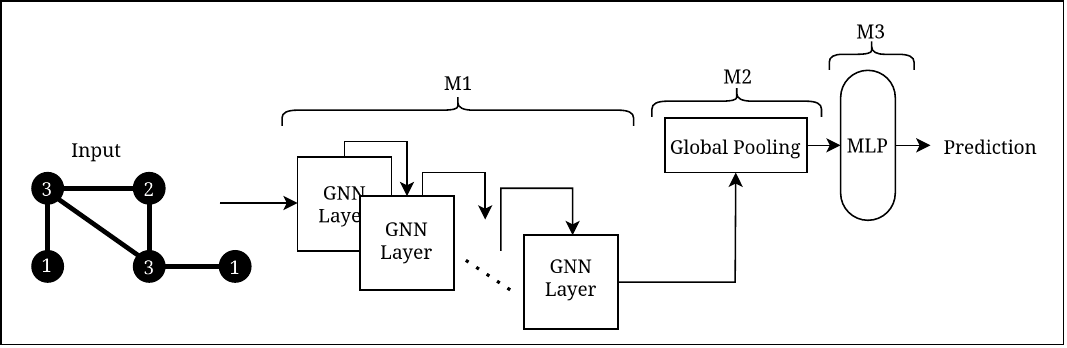}
    \caption{Illustration of the base model $\mathcal{B}$ divided in three modules, M1, M2 and M3. }
    \label{fig:base-model}
\end{figure}

The model ($\mathcal{B}$) consists of three modules. The first module consists of $K$ layers of a GNN. The job of the first module is to work on the graph data and adjust the node embeddings so that the second module, a global pooling function, can summarise them into a single graph-level embedding. The third module is an MLP that takes as input the graph embedding and will adapt it to output the final prediction for the normalised Z-scores of the graphs. 
Figure~\ref{fig:base-model} shows a diagram of the model.

All the optimisations for the hyperparameters of $\mathcal{B}$ were performed by Optuna~\cite{optuna2019} with 450 rounds of suggestions of hyperparameters, orchestrated through Ray~\cite{raytune2018, moritz2018ray}. 
Moreover, the hyperparameter sampling procedure employed the Tree-Structured Parzen Estimator~\cite{watanabe2023treestructured}, while the pruning strategy was executed through the application of the median rule~\cite{Vizier2017medianpruner}.
Table~\ref{tbl:hyper-space} presents the hyperparameter space used for model $\mathcal{B}$.

\begin{table}
\centering
\caption{Break down of the hyperparameter space used for $\mathcal{B}$.}
\label{tbl:hyper-space}
\begin{tabular}{lcc|ccc} 
\toprule
                       & Min & Max                           & Epochs               & Batch Size                                                                         & Learning Rate (log)                 \\ 
\midrule
GNN Depth (M1)         & 2   & 3                             & \multirow{8}{*}{100} & \multirow{8}{*}{\begin{tabular}[c]{@{}c@{}}\{16, 32, 64, \\128, 25\}\end{tabular}} & \multirow{8}{*}{{[}0.00001,0.001]}  \\
Hidden Dimension (M1)  & 6   & 16                            &                      &                                                                                    &                                     \\
GNN Dropout (M1)       & 0.0 & 0.9                           &                      &                                                                                    &                                     \\
Jumping Knowledge (M1) & \multicolumn{2}{c|}{max, cat, lstm} &                      &                                                                                    &                                     \\
Global Pool (M2)       & \multicolumn{2}{c|}{add}            &                      &                                                                                    &                                     \\
MLP Depth (M3)         & 2   & 6                             &                      &                                                                                    &                                     \\
Hidden Dimension (M3)  & 6   & 16                            &                      &                                                                                    &                                     \\
MLP Dropout (M3)       & 0.2 & 0.9                           &                      &                                                                                    &                                     \\
\bottomrule
\end{tabular}
\end{table}

The asymmetry in the hyperparameter space presented in Table~\ref{tbl:hyper-space} stems from our choice of preemptively test a slightly larger hyperparameter space and identify some values that resulted in very bad results. From this early testing phase, we also narrowed down M3 from a global add, mean or max function to just the global add function. This result aligns with some limitations that are known for the mean and max pooling functions~\cite{Xu2019}. AS for the number of epochs, we verified that 100 epochs are generally enough to obtain results before convergence or near convergence of the model. Furthermore, the fixed values of 100 epochs can be shortened not only by the pruner but also by an early-stopping module with a grace period of 25 epochs, synced with the median pruner, and patience of other 25 epochs of not seeing an improvement for the global minimum loss. Moreover, since we believe our problem does not need very long range dependencies since the structures in $\Omega$ can be fully defined by a hop size of 2, in order to try to limit the problem of over-smoothing we limited the maximum number of GNN layers to $3$ based on the findings that most networks have a small diameter~\cite{Albert1999, Barabsi2000, Watts1998}. By limiting the GNN layers, we also hope to reduce over-squashing.

Each experimental iteration with different M1 backbones comprised 450 trials, each uniquely characterised by a distinct combination of hyperparameters suggested by Optuna. 
Most of the training was done using a single consumer-grade NVIDIA RTX 3090 and later a NVIDIA RTX A6000. Inference was performed using a consumer-grade NVIDIA RTX 4070. Computations involving G-trie were performed on a AMD Opteron 6380. 

The code is available in this \href{https://www.github.com/PedrV/motifs-estimation-with-gnns}{Github repository} and the data, figures and other intermediary results are available in this \href{https://figshare.com/s/794d3e3dc66ee09c0e86}{Figshare link}. You can also view the results of the training procedure using the main \href{https://wandb.ai/pedrv/BRONZE}{Wandb} page for this project.


\section{Experiment Results}\label{sec:exp-res-appx}

\begin{figure}[ht]
	\centering
	\includegraphics[clip, width=0.9\textwidth]{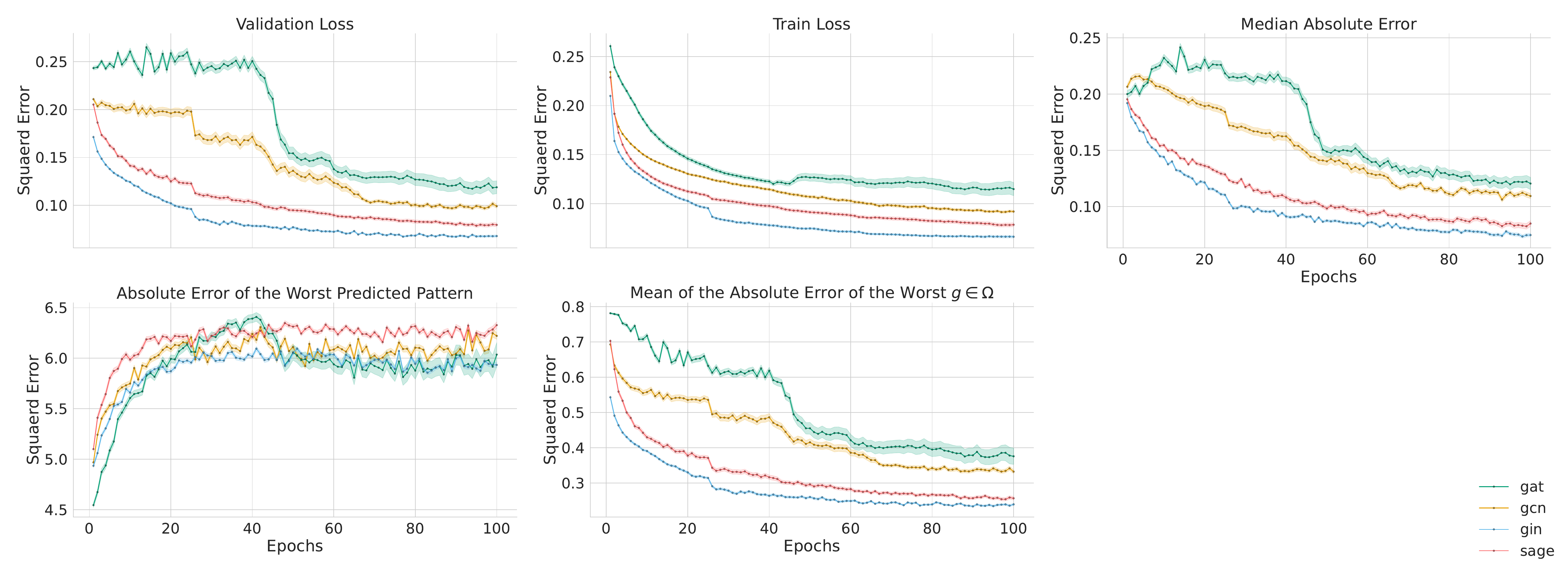}
	\caption{Learning curves for the various backends used for M1 when trained with the deterministic segment of graph generators. }
	\label{fig:learning-curves-d}
\end{figure}

\begin{figure}[ht]
	\centering
	\includegraphics[clip, width=0.9\textwidth]{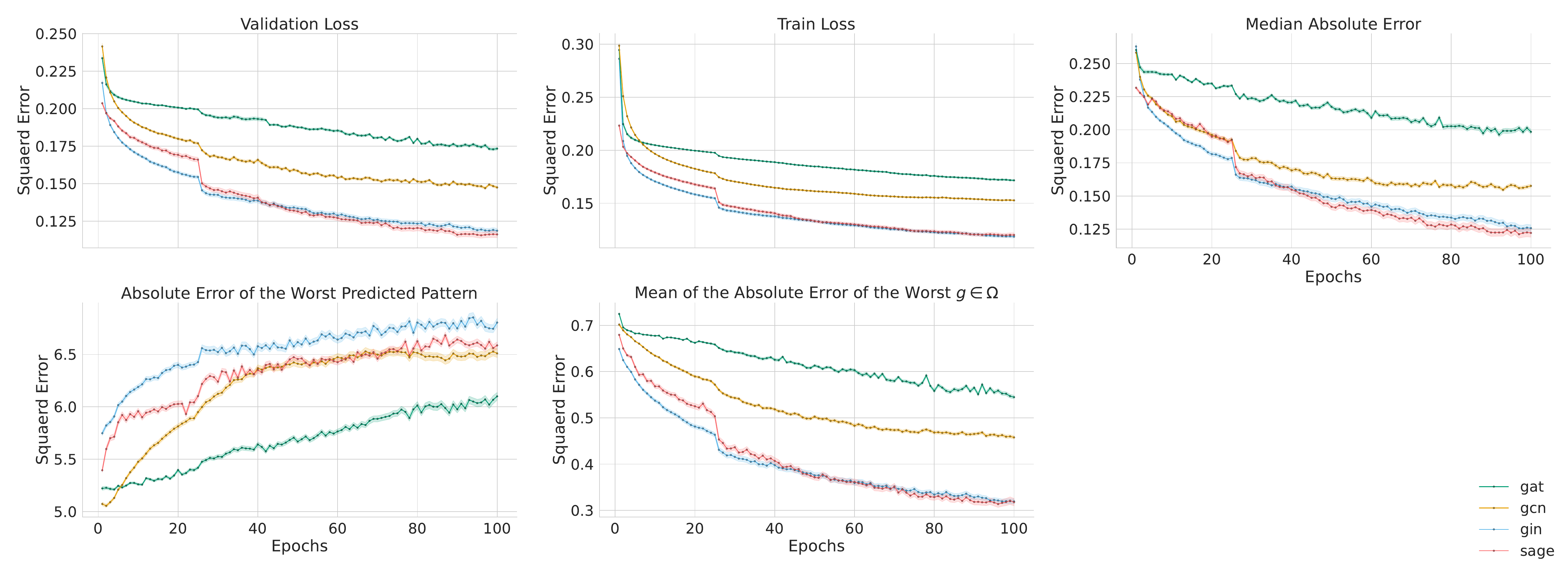}
	\caption{Learning curves for the various backends used for M1 when trained with the non-deterministic segment of graph generators. }
	\label{fig:learning-curves-nd}
\end{figure}

Figure~\ref{fig:learning-curves-d} and~\ref{fig:learning-curves-nd} shows the summary of the results from the 450 rounds of hyperparameter optimisation for each model used in M1. The solid line represents the mean score, and the semi-transparent bound around each line represents the standard error. 
The displayed metrics are the MSE for the train and validation data, the median absolute error, $med \big( \{med_i(|\evy_i - \hat{\evy}_i|, \forall i \in |\vy|) \} \big)$, the maximum absolute error calculated for a full prediction of a significance profile, and the mean value for the worst-performing prediction of a graph from $\Omega$. The maximum error is given by $max( \, \{\sum_{j \in |\Omega|} |\mY_{[i, j]} - \hat{\mY}_{[i,j]}|, \forall i \in |D_\text{valid}| \} \,)$ where $\mY$ is a 2-d matrix with the first dimension giving the number of examples in the validation dataset and the second dimension the length of $\vs$. As for the mean value of the worst-performing predictions, it is given by $mean( \, max\{\sum_{i \in |D_\text{valid}|} |(\mY_{[i, j]} - \hat{\mY}_{[i,j]})|, \forall j \in |\Omega| \} \,)$.

The learning curves for the deterministic segment (Figure~\ref{fig:learning-curves-d}) show that all models improve significantly within the first 50 epochs, especially in all metrics except for the maximum absolute error. GIN outperforms all other models by a wide margin, prompting its selection for further analysis.
For the non-deterministic segment (Figure~\ref{fig:learning-curves-nd}), the performance of GraphSAGE and GIN is very close, with GraphSAGE holding a slight numerical edge. Since both models perform comparably, both will be retained for further evaluation.

\subsection{Predictions}\label{sub:exp-res-preds}

\begin{figure}[htp]

	\begin{subfigure}{\textwidth}
		\includegraphics[clip, clip,width=\textwidth]{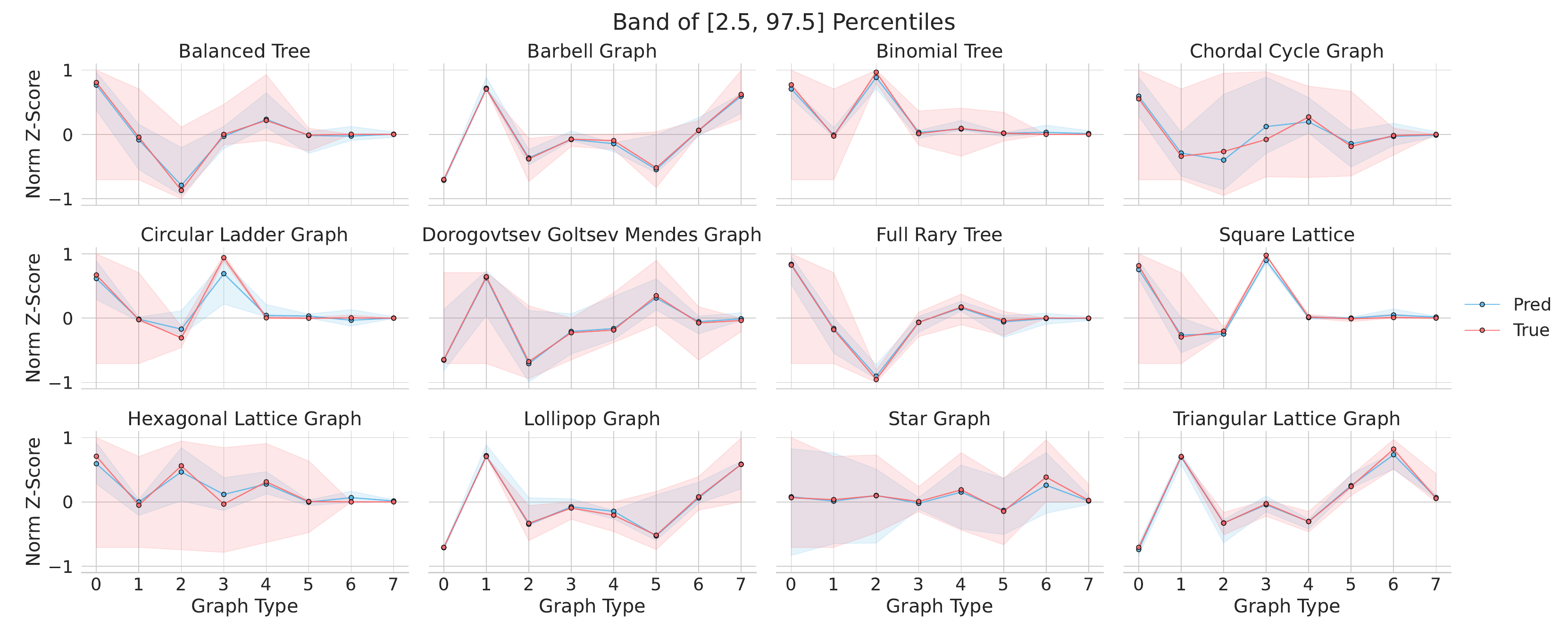}
		\caption{GIN trained on the deterministic segment.}
		\label{fig:synt-dgin-preds}
	\end{subfigure}

    \medskip

	\begin{subfigure}{\textwidth}
		\includegraphics[clip, clip,width=\textwidth]{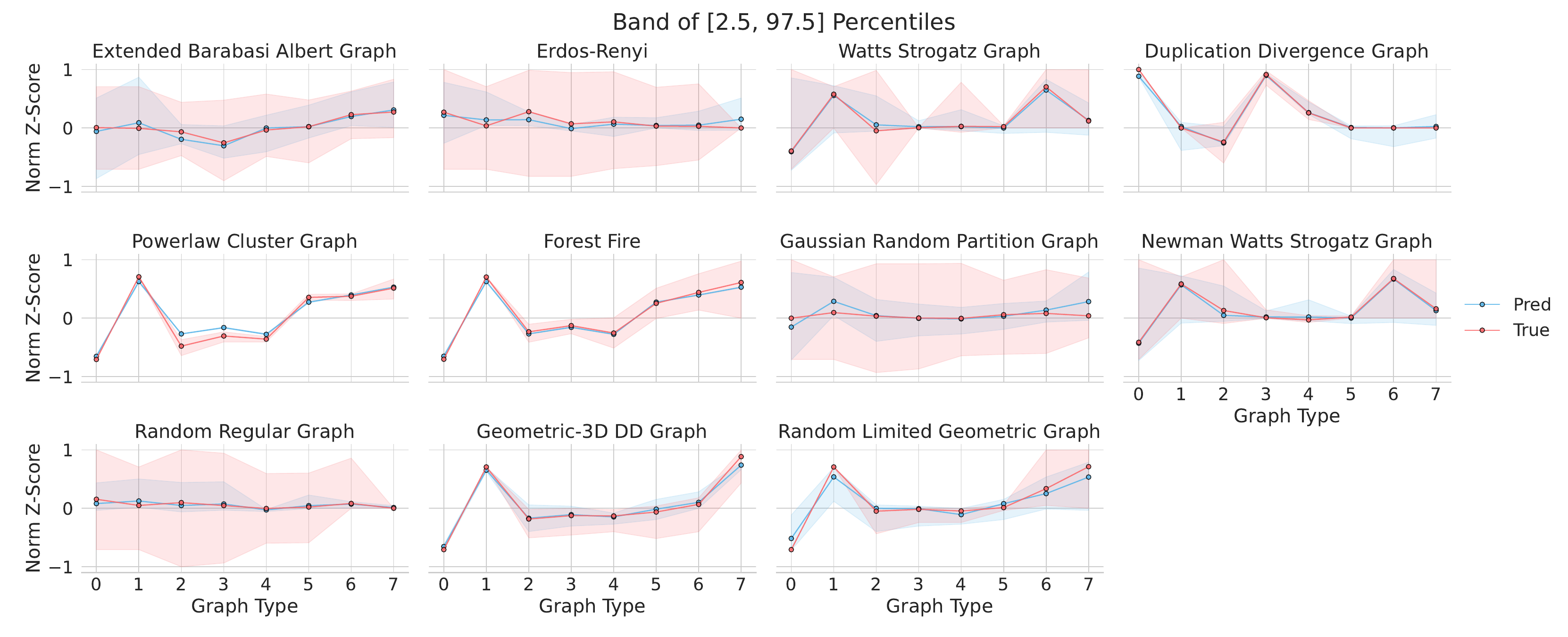}
		\caption{GIN trained on the non-deterministic segment.}
		\label{fig:synt-gin-preds}
	\end{subfigure}

    \medskip

    \begin{subfigure}{\textwidth}
		\includegraphics[clip, clip,width=\textwidth]{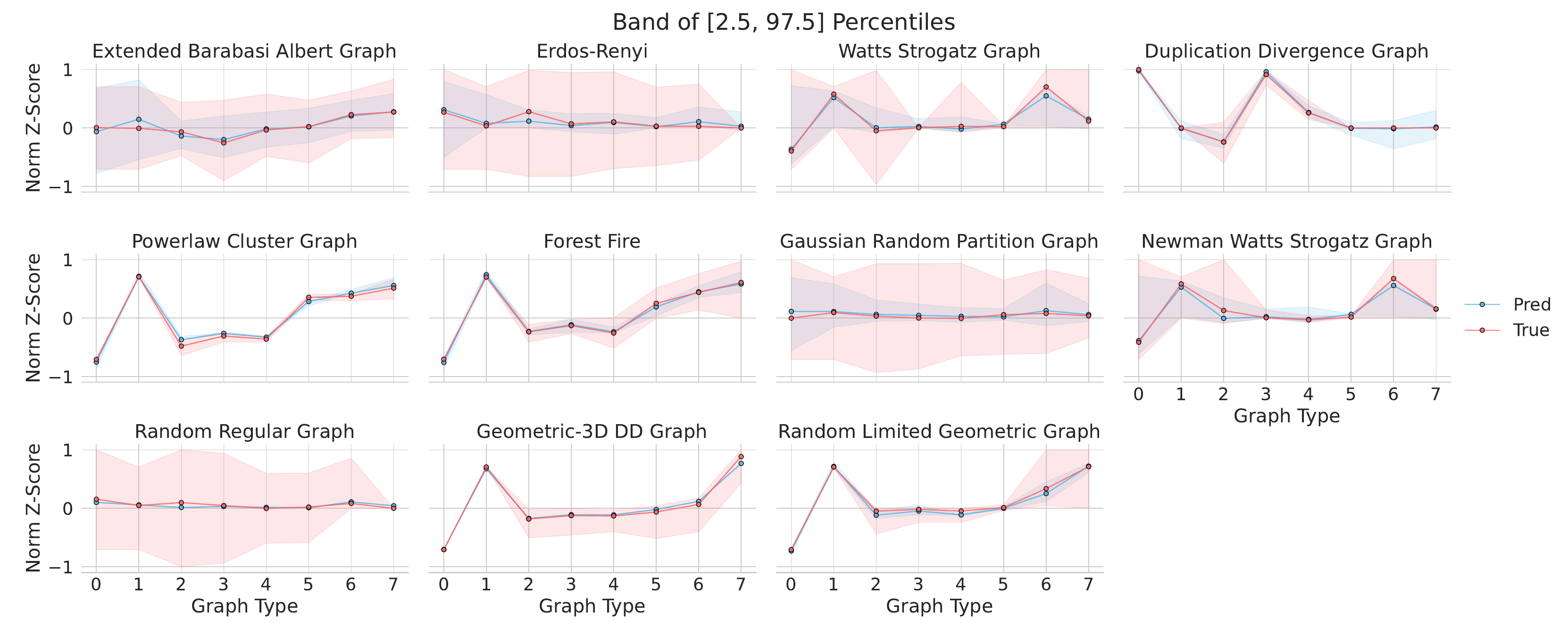}
		\caption{SAGE trained on the non-deterministic segment.}
		\label{fig:synt-sage-preds}
	\end{subfigure}

	\caption{Predictions for each model in each of their corresponding synthetic test datasets. }
    \label{fig:synt-preds}
\end{figure}

Figures \ref{fig:synt-preds} display a summary of the predictions for each generator made by each selected model. The agreement between the true and predicted mean significance-profile is further evidence that the models can perform inter-generator prediction. Furthermore, the much tighter percentile band for the predicted mean significance-profile can be a rough indicator that the models struggle to make accurate intra-generator predictions.

\begin{figure}[htp]

	\begin{subfigure}[ht]{\textwidth}
		\centering
        \includegraphics[width=1\linewidth, align=c]{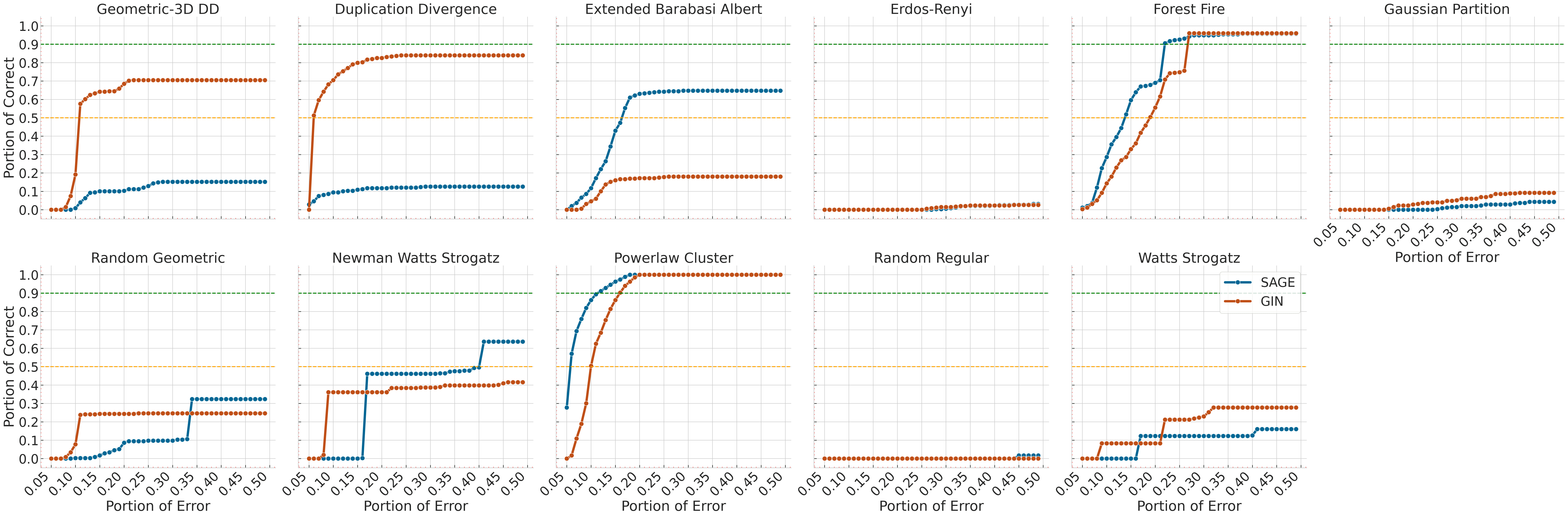}
		\caption{Non-Deterministic Segment}
		\label{fig:correct-evolution-nd} 
	\end{subfigure}

	\smallskip

	\begin{subfigure}[ht]{\textwidth}
		\centering
		\includegraphics[width=1\linewidth, align=c]{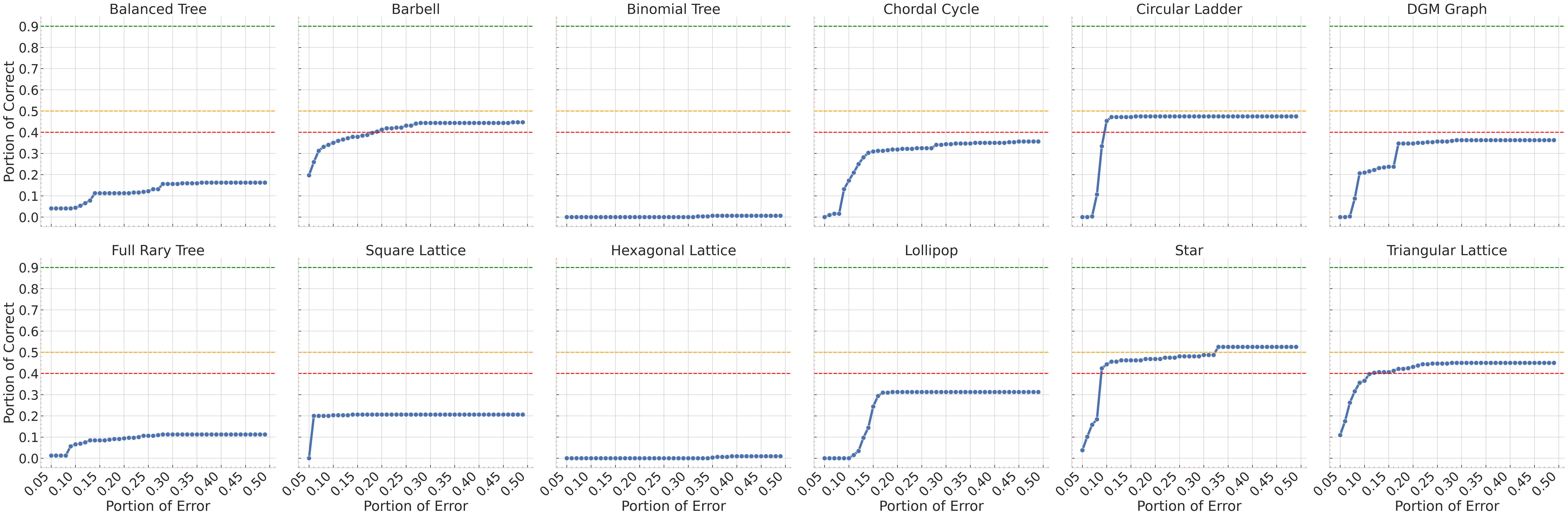}
		\caption{Deterministic Segment}
		\label{fig:correct-evolution-d} 
	\end{subfigure}
	
	\caption{Evolution of the percentage of ``correct'' predictions, as defined in the main text, starting from $5\%$ to $50\%$ in $1\%$. The green line denotes the $90\%$ threshold, the orange the $50\%$ and the red $40\%$.}
\end{figure}

Figure~\ref{fig:correct-evolution-nd} and~\ref{fig:correct-evolution-d} shows a analysis similar to the one made in the main text in tables 1 and 2. In this case, we calculate the percentage of ``correct'' predictions for error thresholds ranging from $5\%$ to $50\%$ in steps of $1\%$. This allows us to see the how volume of ``correct'' predictions evolves over increasing thresholds.

\begin{figure}[htp]

	\begin{subfigure}[ht]{\textwidth}
		\centering
		\includegraphics[width=0.85\linewidth, align=c]{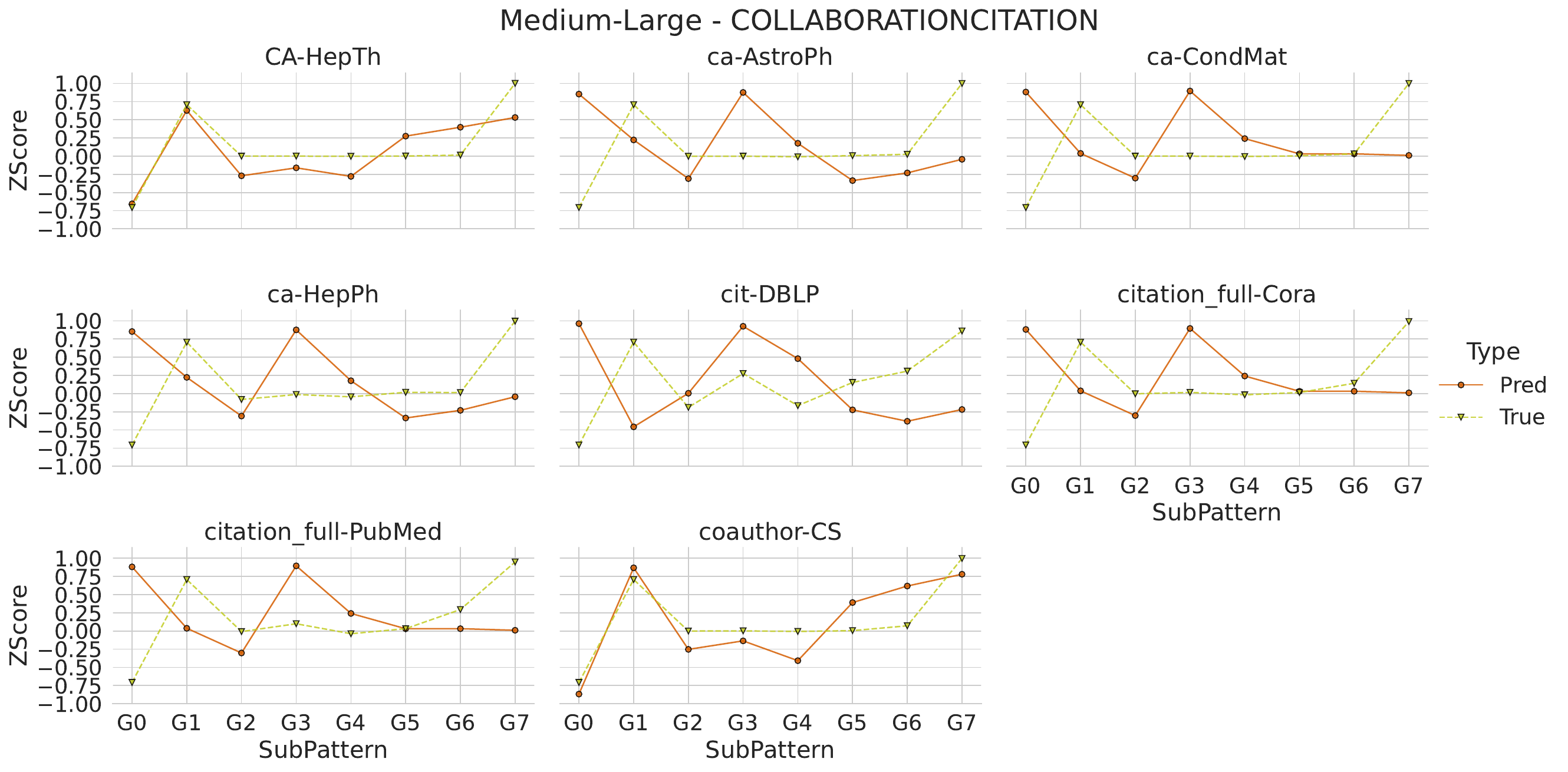}
		\caption{}
		\label{fig:mlreal-colabcit-nd-gin} 
	\end{subfigure}

	\smallskip

	\begin{subfigure}[ht]{\textwidth}
		\centering
		\includegraphics[width=0.85\linewidth, align=c]{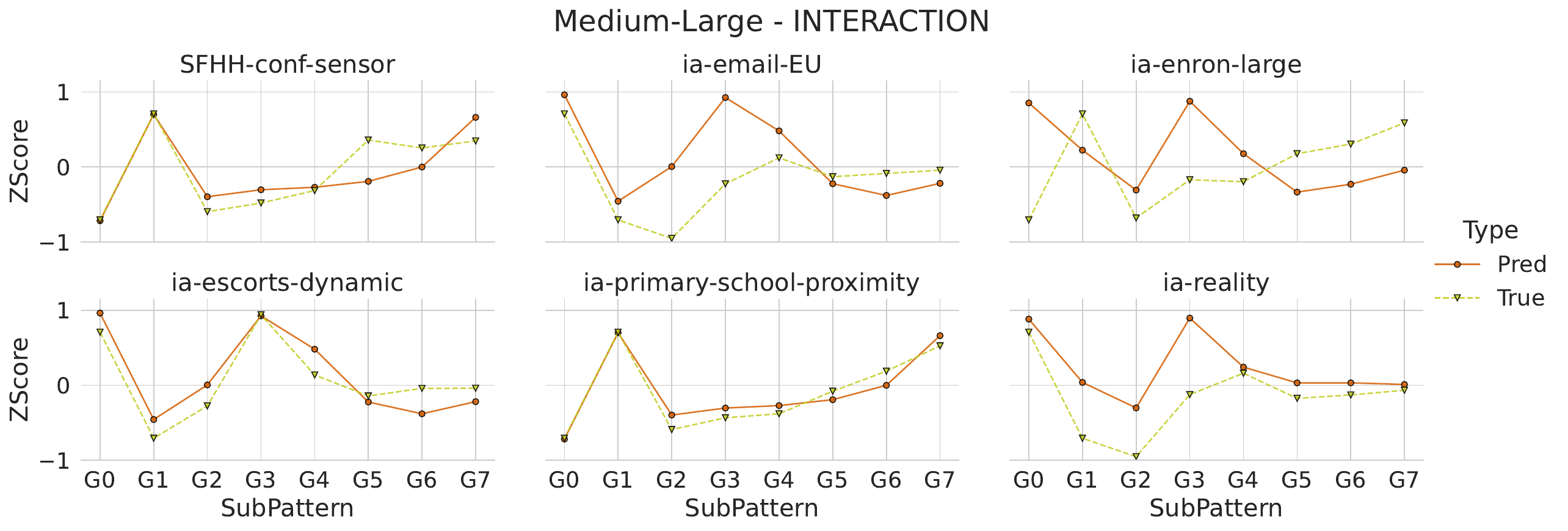}
		\caption{}
		\label{fig:mlreal-inter-nd-gin} 
	\end{subfigure}
	
	\caption{Predictions by GIN trained on the non-deterministic segment. Orange lines with circles are predictions and dark-yellow with triangles true values. }
\end{figure}

Figure~\ref{fig:mlreal-inter-nd-gin} and~\ref{fig:mlreal-colabcit-nd-gin} show some examples of predictions made in the real-world dataset. We can see that even though the model generally struggles to make accurate intra-generator predictions, it makes predictions that can be traced back to the synthetic generator. The \textit{ia-escorts-dynamic}, \textit{coauthor-CS}, and \textit{ia-primary-school-proximity} are the examples highlighted in the main text. For these specific examples, the significance-profiles can be traced back to the duplication-divergence, forest-fire, and geometric generators respectively. This suggests that these generators produce the most similar graphs to real-world networks from a motif analysis perspective. This alignment is expected, as these generators are designed to replicate such patterns.


\end{document}